\documentclass[11pt]{article}
\usepackage{amsmath, amssymb, amsthm, amsfonts}
\usepackage{bm}
\usepackage{graphicx}
\usepackage[table]{xcolor}
\usepackage{multirow}
\usepackage{algorithm}
\usepackage{algpseudocode}
\usepackage{booktabs}
\usepackage{tabularx}
\usepackage{makecell}
\usepackage[section]{placeins} 
\graphicspath{{figures/cls_vis/}{figures/seg_vis/}}
\usepackage{placeins}

\usepackage[left=1in, right=1in, top=0.8in, bottom=1in, 
            headsep=10pt, textwidth=6.65in, textheight=8.7in]{geometry}

\usepackage{authblk}

\title{\textbf{Multimodal Structure Learning: Disentangling Shared and Specific Topology via Cross-Modal Graphical Lasso}}

\author[1]{Fei Wang$^{*}$}
\author[2]{Yutong Zhang$^{*}$}
\author[3]{Xiong Wang$^{\dagger}$}

\affil[1]{Department of Applied Mathematics and Statistics, Stony Brook University \\ 
         \texttt{fei.wang.1@stonybrook.edu}}
\affil[2]{School of Mathematics, Sichuan University \\ \texttt{yutongzhang@stu.scu.edu.cn}} 
\affil[3]{School of Computer Science and Technology, USTC \\ 
         \texttt{wangxiong@ustc.edu.cn}}

\date{}

\begin{document}
\renewcommand{\arraystretch}{1.3}  

\maketitle
\vspace{-0.5cm}
\footnotetext[1]{$^{*}$Equal contribution.}
\footnotetext[2]{$^{\dagger}$Corresponding author.}
\begin{abstract}
Learning interpretable multimodal representations inherently relies on uncovering the conditional dependencies between heterogeneous features. However, sparse graph estimation techniques, such as Graphical Lasso (GLasso), to visual-linguistic domains is severely bottlenecked by high-dimensional noise, modality misalignment, and the confounding of shared versus category-specific topologies. In this paper, we propose Cross-Modal Graphical Lasso (CM-GLasso) that overcomes these fundamental limitations. By coupling a novel text-visualization strategy with a unified vision-language encoder, we strictly align multimodal features into a shared latent space. We introduce a cross-attention distillation mechanism that condenses high-dimensional patches into explicit semantic nodes, naturally extracting spatial-aware cross-modal priors. Furthermore, we unify tailored GLasso estimation and Common-Specific Structure Learning (CSSL) into a joint objective optimized via the Alternating Direction Method of Multiplier (ADMM). This formulation guarantees the simultaneous disentanglement of invariant and class-specific precision matrices without multi-step error accumulation. Extensive experiments across eight benchmarks covering both natural and medical domains demonstrate that CM-GLasso establishes a new state-of-the-art in generative classification and dense semantic segmentation tasks.
\end{abstract}

\section{Introduction}
\label{sec:intro}

Discovering conditional dependency structures among multimodal features is fundamental for interpretable representation learning. While Graphical Lasso (GLasso)~\cite{friedman2008sparse} remains the dominant approach for estimating sparse precision matrices in Gaussian Graphical Models (GGMs), applying it to multimodal visual scenarios reveals three critical challenges:

\textbf{First, the high-dimensional low-sample-size (HDLSS) problem.} Deep visual features ($p \gg n$) make empirical covariance matrices highly unreliable. Standard GLasso's uniform $\ell_1$ penalization struggles to distinguish genuine conditional dependencies from dense spurious edges.

\textbf{Second, insufficient cross-modal exploitation.} Existing methods process modalities independently or via simple concatenation, failing to utilize the structural prior of one modality (e.g., high-level text semantics) to guide the graph estimation of another (e.g., low-level visual patterns).

\textbf{Third, ignoring shared-versus-specific topological structures.} Estimating graphs independently per category discards invariant shared patterns (e.g., foreground-background separation) while failing to isolate category-specific structural nuances.

While approaches like Tailored GLasso~\cite{lingjarde2021tailored} successfully leverage auxiliary priors via an eBIC-guided sigmoid transformation, they remain confined to bioinformatics and unimodal settings, leaving the multimodal visual-linguistic domain unexplored.

To address these limitations, we propose \textbf{CM-GLasso} (Cross-Modal Graphical Lasso) (Figure~\ref{fig:pipeline}), a framework guided by four key insights:

\textbf{1. Unified Representation \& Prior Transfer:} We introduce a \textit{text visualization} strategy, encoding both text (rendered as images) and actual images through a single vision-language encoder (SigLIP~2 ViT~\cite{tschannen2025siglip2}). This guarantees that cross-modal features reside in a shared embedding space with naturally aligned attention structures.

\textbf{2. Cross-Attention Distillation:} Instead of blind dimensionality reduction (e.g., PCA or FC layers), we condense $N_p$ patch features into $p$ semantic nodes via learnable prototypes. Their spatial attention co-occurrences naturally formulate a dimensionally-aligned $p \times p$ cross-modal prior matrix.

\textbf{3. Data-Adaptive Prior Utilization:} Auxiliary priors are not universally beneficial. We employ an eBIC-guided mechanism to dynamically control the sigmoid sharpness parameter $k^*$, allowing the framework to gracefully degrade to standard GLasso ($k^* = 0$) when priors are uninformative.

\textbf{4. Joint Optimization:} We unify tailored GLasso estimation and Common-Specific Structure Learning (CSSL) into a single objective solved via ADMM~\cite{boyd2011distributed}, preventing the error accumulation inherent in two-stage decompositions.

Our main contributions are summarized as follows:
\vspace{-2mm}
\begin{itemize}
    \item We propose a text visualization strategy and cross-attention distillation mechanism that seamlessly resolve feature space inconsistencies and automatically extract highly interpretable cross-modal structural priors.
    \item We formulate a joint objective unifying Tailored GLasso and CSSL, optimized via ADMM, for the end-to-end disentanglement of shared and category-specific graph topologies.
    \item We design unified task-specific heads (classification $\mathcal{H}_C$ and segmentation $\mathcal{H}_S$) over the shared structural representation, demonstrating significant improvements across eight benchmarks and effectively extending tailored GLasso into the multimodal visual domain.
\end{itemize}
\section{Related Work}
\label{sec:related}

\textbf{Sparse Precision Matrix Estimation.} 
Graphical Lasso (GLasso) \cite{friedman2008sparse} and its variants have long been the gold standard for $\ell_1$-regularized precision matrix estimation. Subsequent advancements, such as non-uniform penalty weighting \cite{ambroise2009inferring} and eBIC-guided structure selection \cite{foygel2010extended, lingjarde2021tailored}, have further enhanced estimation robustness. In the visual domain, Souly and Shah \cite{souly2016scene} demonstrated the efficacy of sparse precision matrices in capturing long-range label interactions for scene labeling. However, these classical statistical approaches are predominantly confined to unimodal data. They lack the mechanism to construct and inject cross-modal topological priors, and critically, they treat graph estimation and topology decomposition as decoupled, multi-step processes.

\textbf{Vision-Language Representation Learning.} 
Recent foundation models, from CLIP \cite{radford2021learning} to the state-of-the-art SigLIP 2 \cite{tschannen2025siglip2}, have established highly aligned cross-modal embedding spaces. While recent unified architectures like PaliGemma 2 \cite{steiner2024paligemma2} and Ja \cite{wu2025ja} leverage these encoders for multimodal understanding and generation, their application to conditional dependency modeling remains underexplored. Specifically, how to exploit the shared geometric pathways of unified encoders (e.g., spatial co-occurrence of attention footprints) to transfer graph structural priors across modalities is an open problem. Our CM-GLasso bridges this exact gap by unifying vision-language pretraining priors with statistical graph estimation in a unified, topology-aware framework.
\section{Methodology}
\label{sec:method}

\begin{figure*}[t]
    \centering
    \includegraphics[width=\textwidth]{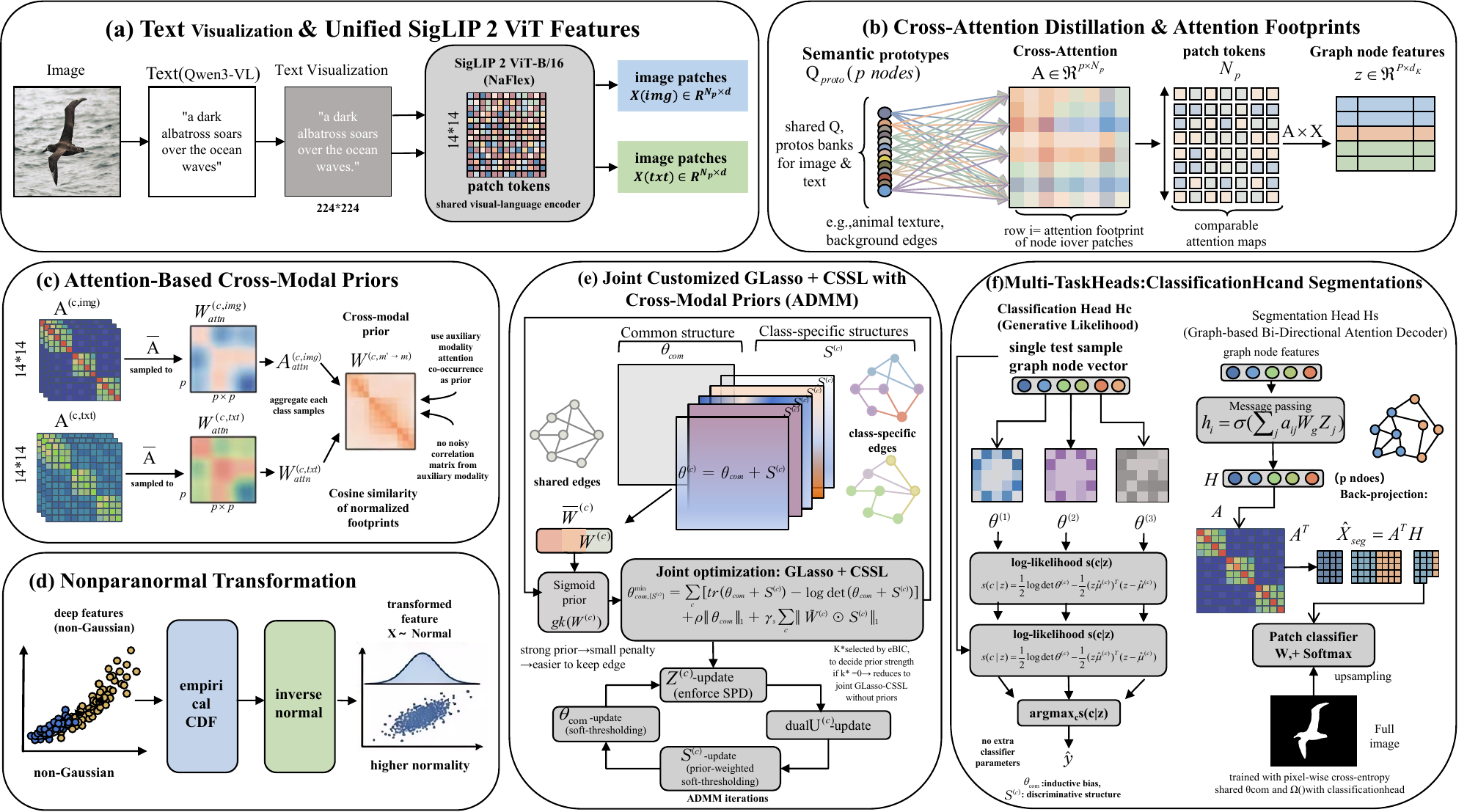}
    \caption{The Pipeline of CM-GLasso. (a) Text visualization and unified SigLIP 2 feature extraction. (b) Cross-attention distillation condenses patches into semantic nodes. (c) Attention footprints derive spatial-aware cross-modal priors. (d-e) Nonparanormal transformation strictly ensures Gaussianity for the subsequent joint ADMM optimization, simultaneously disentangling shared ($\boldsymbol{\Theta}_{\text{com}}$) and specific ($\boldsymbol{S}^{(c)}$) topologies. (f) Learned structures govern generative classification and topology-aware segmentation.}
    \label{fig:pipeline}
\end{figure*}

Figure~\ref{fig:pipeline} illustrates the overall pipeline of the CM-GLasso framework. Given an image-text dataset $\mathcal{D}$ with $C$ categories, our framework operates in three stages: (1) \textbf{Cross-Modal Prior Construction}: mapping heterogeneous inputs into a unified latent space via SigLIP~2~\cite{tschannen2025siglip2}, condensing graph nodes via cross-attention distillation, and extracting spatial-aware priors from attention footprints; (2) \textbf{Prior-Guided Structure Learning}: applying nonparanormal transformation~\cite{liu2009nonparanormal} to satisfy Gaussianity, then jointly optimizing common and class-specific precision matrices via ADMM~\cite{boyd2011distributed}; (3) \textbf{Graph-Structured Inference}: leveraging learned topologies for classification via likelihood-based discrimination and segmentation via topology-aware message passing.

\subsection{Problem Formulation}

Let $\mathbf{x}_n^{(m)} \in \mathbb{R}^{p}$ denote the feature vector of the $n$-th sample under modality $m$, and $y_n \in \{1, \ldots, C\}$ be the corresponding class label. For each class $c$ and modality $m$, our objective is to estimate the precision matrix $\boldsymbol{\Theta}^{(c,m)} = (\boldsymbol{\Sigma}^{(c,m)})^{-1}$ from $n_c$ samples, and decompose it into a common structure $\boldsymbol{\Theta}_{\text{com}}$ and a class-specific structure $\boldsymbol{S}^{(c)}$, following the joint graphical lasso framework~\cite{danaher2014joint}. A non-zero off-diagonal element $\theta_{ij} \neq 0$ in the precision matrix implies a \textit{conditional dependence} between features $i$ and $j$ (where $i,j \in \{1,\ldots,p\}$ denote feature/node indices). 

Statistically, the exact partial correlation is given by $\rho_{ij|V\setminus\{i,j\}} = -\theta_{ij} / \sqrt{\theta_{ii}\theta_{jj}}$. Consequently, the sign of $\theta_{ij}$ is strictly opposite to the actual conditional dependence: $\theta_{ij} < 0$ indicates synergistic positive correlation (e.g., texture co-occurrence), while $\theta_{ij} > 0$ implies mutually exclusive negative correlation (e.g., competing semantic roles). Rather than computing explicit partial correlations which introduces division operations, our subsequent graph-structured inference directly partitions the edges based on the sign of $\theta_{ij}$. This mathematically preserves the bipartite physical semantics of the structural pathways while ensuring numerical stability during representation learning.

\subsection{Unified Multimodal Feature Extraction}

\subsubsection{Text Visualization}

A key challenge in multimodal learning is that images and texts typically use encoders with disparate architectures (e.g., ViT vs. BERT), requiring additional alignment modules that operate at the embedding level without guaranteeing consistency at the attention structure level. To address this, we adopt a \textit{Text Visualization} strategy inspired by vision-language pre-training~\cite{radford2021learning}: given a text description $T$, we render it into a $224 \times 224$ image $I_T$ (black text on white background with adaptive font sizing), and extract features using the \textit{same vision-language pre-trained encoder}. This eliminates the cross-modal gap, requires only a single encoder, and avoids the overhead of an auxiliary language model. The efficacy of this strategy hinges on the encoder's capacity to comprehend rendered text, motivating our choice of a vision-language pre-trained model.

\subsubsection{Unified Vision-Language Encoder: SigLIP~2 ViT}

We employ the ViT-B/16 architecture of SigLIP~2~\cite{tschannen2025siglip2} as our unified multimodal feature extractor. Jointly pre-trained on vision-language data with a sigmoid loss~\cite{zhai2023sigmoid}, SigLIP~2 ensures a high-quality shared latent space. Its \textbf{NaFlex variant} supports multi-resolution inputs while preserving aspect ratios, making it well-suited for rendered text images.

Given an input image (visual or rendered text) $I \in \mathbb{R}^{H \times W \times 3}$, the final layer (excluding [CLS] token) outputs a feature matrix $\mathbf{X}^{(m)} \in \mathbb{R}^{N_p \times d}$ with $N_p = 196$ patches and $d = 768$, where $m \in \{\text{img}, \text{txt}\}$. Both modalities share the identical encoder and weights, ensuring their patch features reside in the same embedding space.

\subsubsection{Cross-Attention Distillation: From Patches to Graph Nodes}
\label{sec:cross_attn_distill}

Graphical models require $p$ nodes ($p \ll N_p$) to compute a $p \times p$ precision matrix. We introduce \textit{cross-attention distillation} to distill $N_p$ unordered patches into $p$ semantic graph nodes with explicit origins. We declare $p$ learnable semantic prototypes $\mathbf{Q}_{\text{proto}} \in \mathbb{R}^{p \times d}$, which converge during training into semantic probes (e.g., probe 1 capturing ``animal textures,'' probe 2 capturing ``background edges'').

Using linear projections $\mathbf{W}_Q, \mathbf{W}_K, \mathbf{W}_V \in \mathbb{R}^{d \times d_k}$, we construct standard cross-attention. For a given sample, the cross-attention matrix is:
\begin{equation}
\label{eq:cross_attn}
\mathbf{A} = \text{softmax}\left(\frac{(\mathbf{Q}_{\text{proto}}\mathbf{W}_Q)(\mathbf{X}\mathbf{W}_K)^\top}{\sqrt{d_k}}\right) \in \mathbb{R}^{p \times N_p}
\end{equation}
The $i$-th row $\mathbf{A}_{i,:}$ records attention weights of the $i$-th semantic probe across the $N_p$ patches. Graph node features for this sample are extracted as:
\begin{equation}
\label{eq:node_feature_matrix}
\mathbf{Z} = \mathbf{A} (\mathbf{X} \mathbf{W}_V) \in \mathbb{R}^{p \times d_k}
\end{equation}
For the $n$-th sample, we denote its node feature matrix as $\mathbf{Z}_n \in \mathbb{R}^{p \times d_k}$. To adapt to the GLasso's Gaussian input space, we aggregate along the feature channel dimension via a learnable projection $\mathbf{w}_{\text{out}} \in \mathbb{R}^{d_k}$, yielding a $p$-dimensional observation vector:
\begin{equation}
\label{eq:node_vector}
\mathbf{z}_n = \mathbf{Z}_n \mathbf{w}_{\text{out}} \in \mathbb{R}^p
\end{equation}
Since $\mathbf{Q}_{\text{proto}}$ can be shared across modalities, the resulting attention matrices $\mathbf{A}^{(\text{img})}$ and $\mathbf{A}^{(\text{txt})}$ are inherently comparable.

\subsubsection{Construction of the Prior Matrix}
\label{sec:prior_from_attn}

The cross-attention matrix $\mathbf{A}$ also provides a bridge for constructing the $p \times p$ prior matrix. If the footprints of graph nodes $i$ and $j$ (rows of $\mathbf{A}$) highly overlap—attending to similar image patches—they likely share conditional dependence. We first perform $\ell_2$-normalization on the aggregated attention distributions:
\begin{equation}
\label{eq:attn_norm}
\bar{\mathbf{A}}_{i,:}^{(c,m)} = \frac{\mathbf{A}_{\text{agg}, i,:}^{(c,m)}}{\|\mathbf{A}_{\text{agg}, i,:}^{(c,m)}\|_2}
\end{equation}
We then define prior weights via cosine similarity:
\begin{equation}
\label{eq:prior_attn}
\mathbf{W}_{\text{attn}}^{(c, m)} = \bar{\mathbf{A}}^{(c,m)} (\bar{\mathbf{A}}^{(c,m)})^\top \in \mathbb{R}^{p \times p}
\end{equation}
The cross-modal prior is given by the attention co-occurrence matrix of the auxiliary modality:
\begin{equation}
\label{eq:cross_prior}
\mathbf{W}^{(c, m' \to m)} = \mathbf{W}_{\text{attn}}^{(c, m')}
\end{equation}
We use only the auxiliary modality's attention co-occurrence to guide the target modality's graph structure. Under the HDLSS regime, empirical correlation matrices are noisy; attention footprints reflect topological spatial co-occurrence in a strictly aligned cross-modal space, offering greater structural reliability.

\subsection{Nonparanormal Transformation}
\label{sec:nonparanormal}

GLasso assumes multivariate normality, but Transformer features are typically non-Gaussian (only $\sim$23\% of dimensions pass the Shapiro-Wilk test). We apply the nonparanormal transformation~\cite{liu2009nonparanormal}. For the $j$-th dimension of $\mathbf{z}_n$:
\begin{equation}
\label{eq:nonparanormal}
\tilde{z}_{nj} = \Phi^{-1}\left(\hat{F}_j(z_{nj})\right), \quad j = 1, \ldots, p
\end{equation}
where $\hat{F}_j$ is the empirical CDF and $\Phi^{-1}$ is the standard normal quantile function. Following Liu et al.~\cite{liu2009nonparanormal}, we use a rank-based empirical CDF:
\begin{equation}
\label{eq:empirical_cdf}
\hat{F}_j(z_{nj}) = \frac{\text{rank}(z_{nj}) - 0.5}{n_c}
\end{equation}
After transformation, the normality test pass rate improves from $\sim$23\% to $\sim$88\% (see ablation studies in Sec.~\ref{sec:ablation}).

Crucially, the transformed features $\tilde{\mathbf{z}}_n = [\tilde{z}_{n1}, \ldots, \tilde{z}_{np}]^\top$ naturally adhere to a standard normal distribution with zero mean. Thus, for samples belonging to class $c$, the class-conditional empirical covariance matrix is rigorously formulated as:
\begin{equation}
\label{eq:empirical_cov}
\hat{\boldsymbol{\Sigma}}^{(c)} = \frac{1}{n_c}\sum_{n: y_n = c} \tilde{\mathbf{z}}_n \tilde{\mathbf{z}}_n^\top
\end{equation}
This robust covariance estimator $\hat{\boldsymbol{\Sigma}}^{(c)}$ is subsequently utilized as the data-driven input for our topological optimization.

\subsection{Unified Optimization of Tailored GLasso and CSSL}
\label{sec:joint_opt}

Traditional approaches estimate the precision matrix via tailored GLasso~\cite{lingjarde2021tailored} and then decompose it via CSSL~\cite{danaher2014joint} in two separate steps, leading to error accumulation. We propose a unified framework that jointly optimizes $\boldsymbol{\Theta}_{\text{com}}$ and $\{\boldsymbol{S}^{(c)}\}_{c=1}^C$, inspired by joint graphical lasso~\cite{danaher2014joint} and common substructure learning~\cite{hara2013learning}:
\begin{equation}
\label{eq:joint}
\begin{aligned}
\min_{\boldsymbol{\Theta}_{\text{com}}, \{\boldsymbol{S}^{(c)}\}} \quad & \sum_{c=1}^{C} \Big[ \text{tr}\!\left(\hat{\boldsymbol{\Sigma}}^{(c)}(\boldsymbol{\Theta}_{\text{com}} + \boldsymbol{S}^{(c)})\right) - \log\det(\boldsymbol{\Theta}_{\text{com}} + \boldsymbol{S}^{(c)}) \Big] \\
& + \rho \|\boldsymbol{\Theta}_{\text{com}}\|_1 + \gamma_s \sum_{c=1}^{C} \|\tilde{\mathbf{W}}^{(c)} \odot \boldsymbol{S}^{(c)}\|_1 \\
\text{s.t.} \quad & \boldsymbol{\Theta}_{\text{com}} + \boldsymbol{S}^{(c)} \succ 0, \quad \forall c
\end{aligned}
\end{equation}

The adaptive weight matrix $\tilde{\mathbf{W}}^{(c)} \in \mathbb{R}^{p \times p}$ is defined via a sigmoid transformation of the cross-modal prior $\mathbf{W}^{(c, m' \to m)}$:
\begin{equation}
\label{eq:adaptive_weight}
\tilde{w}_{ij}^{(c)} = 1 - \frac{1}{1 + \exp(-k^*(W_{ij}^{(c, m' \to m)} - 0.5))}
\end{equation}
where $k^*$ is automatically selected via eBIC~\cite{foygel2010extended} to control prior sharpness. This design ensures that when the auxiliary modality indicates strong co-occurrence ($W_{ij}^{(c)} \gg 0.5$), $\tilde{w}_{ij}^{(c)} \to 0$, preserving that edge in $\boldsymbol{S}^{(c)}$; when the prior is weak, $\tilde{w}_{ij}^{(c)} \to 1$, applying full $\ell_1$ regularization. The cross-modal prior is injected specifically into the class-specific structures, enabling the model to preserve semantically meaningful edges (e.g., ``cat+sofa'') in relevant classes.

\subsubsection{Efficient Optimization via ADMM}
\label{sec:admm}

We solve Eq.~\eqref{eq:joint} using the alternating direction method of multipliers (ADMM)~\cite{boyd2011distributed}. Introducing auxiliary variables $\mathbf{Z}^{(c)} = \boldsymbol{\Theta}_{\text{com}} + \boldsymbol{S}^{(c)}$ and dual variables $\mathbf{U}^{(c)}$, the augmented Lagrangian is:
\begin{equation}
\begin{aligned}
\mathcal{L}_{\mu} = & \sum_{c=1}^{C} \Big[ \text{tr}(\hat{\boldsymbol{\Sigma}}^{(c)}\mathbf{Z}^{(c)}) - \log\det\mathbf{Z}^{(c)} + \langle \mathbf{U}^{(c)}, \mathbf{Z}^{(c)} - \boldsymbol{\Theta}_{\text{com}} - \boldsymbol{S}^{(c)} \rangle \\
& + \frac{\mu}{2} \|\mathbf{Z}^{(c)} - \boldsymbol{\Theta}_{\text{com}} - \boldsymbol{S}^{(c)}\|_F^2 \Big] + \rho \|\boldsymbol{\Theta}_{\text{com}}\|_1 + \gamma_s \sum_{c=1}^{C} \|\tilde{\mathbf{W}}^{(c)} \odot \boldsymbol{S}^{(c)}\|_1
\end{aligned}
\end{equation}
The optimization decouples into four subproblems:

\textbf{Update $\mathbf{Z}^{(c)}$}: Solved via eigenvalue decomposition. We minimize a Frobenius norm penalized by a log-determinant barrier:
\begin{equation}
\label{eq:z_update}
\mathbf{Z}^{(c)} \leftarrow \arg\min_{\mathbf{Z} \succ 0} \left\{\text{tr}(\hat{\boldsymbol{\Sigma}}^{(c)}\mathbf{Z}) - \log\det\mathbf{Z} + \frac{\mu}{2}\|\mathbf{Z} - \mathbf{G}^{(c)}\|_F^2\right\}
\end{equation}
where $\mathbf{G}^{(c)} = \boldsymbol{\Theta}_{\text{com}} + \boldsymbol{S}^{(c)} - \mu^{-1}\mathbf{U}^{(c)}$. Specifically, let $\mathbf{Q} \boldsymbol{\Lambda} \mathbf{Q}^\top$ be the eigendecomposition of the symmetric matrix $\mathbf{G}^{(c)} - \mu^{-1}\hat{\boldsymbol{\Sigma}}^{(c)}$, where $\boldsymbol{\Lambda} = \text{diag}(\lambda_1, \dots, \lambda_p)$. The optimal positive-definite update is derived by applying a non-negative soft-thresholding operator to the eigenvalues, yielding $\mathbf{Z}^{(c)} = \mathbf{Q} \tilde{\boldsymbol{\Lambda}} \mathbf{Q}^\top$, with the diagonal elements updated as $\tilde{\lambda}_i = \frac{1}{2}(\lambda_i + \sqrt{\lambda_i^2 + 4/\mu})$.

\textbf{Update $\boldsymbol{\Theta}_{\text{com}}$}: Closed-form soft-thresholding:
\begin{equation}
\label{eq:theta_update}
\boldsymbol{\Theta}_{\text{com}} \leftarrow \mathcal{S}_{\rho/(C\mu)}\!\left(\frac{1}{C}\sum_c (\mathbf{Z}^{(c)} - \boldsymbol{S}^{(c)} + \mu^{-1}\mathbf{U}^{(c)})\right)
\end{equation}

\textbf{Update $\boldsymbol{S}^{(c)}$}: Element-wise soft-thresholding guided by cross-modal prior:
\begin{equation}
\label{eq:omega_update}
\boldsymbol{S}^{(c)} \leftarrow \mathcal{S}_{\gamma_s \tilde{\mathbf{W}}^{(c)}/\mu}\!\left(\mathbf{Z}^{(c)} - \boldsymbol{\Theta}_{\text{com}} + \mu^{-1}\mathbf{U}^{(c)}\right)
\end{equation}

\textbf{Update $\mathbf{U}^{(c)}$}: Dual ascent step:
\begin{equation}
\label{eq:dual_update}
\mathbf{U}^{(c)} \leftarrow \mathbf{U}^{(c)} + \mu(\mathbf{Z}^{(c)} - \boldsymbol{\Theta}_{\text{com}} - \boldsymbol{S}^{(c)})
\end{equation}

The parameter $k^*$ (controlling sigmoid sharpness) is selected via eBIC~\cite{foygel2010extended}. If $k^*=0$, the framework gracefully degrades to prior-free joint optimization. The complete algorithm is summarized in the supplementary materials.

\textbf{Remark on Positive Definiteness for Inference:} In the ADMM formulation, the auxiliary variables $\mathbf{Z}^{(c)}$ are explicitly constrained to be positive definite via the log-determinant subproblem (Eq.~\eqref{eq:z_update}), ensuring $\mathbf{Z}^{(c)} \succ 0$ at every iteration. However, the reconstructed precision matrix $\boldsymbol{\Theta}_{\text{com}} + \boldsymbol{S}^{(c)}$ may not be strictly positive definite until the algorithm reaches convergence, especially under early stopping. To guarantee numerical stability during downstream inference—particularly when computing $\log\det(\boldsymbol{\Theta}^{(c)})$ in the classification head (Eq.~\eqref{eq:cls_head})—we directly substitute the strictly positive definite auxiliary variable as the final precision matrix estimate:
\begin{equation}
\label{eq:inference_precision}
\hat{\boldsymbol{\Theta}}^{(c)} = \mathbf{Z}^{(c)} \quad \text{(after convergence)}
\end{equation}
This substitution is mathematically justified by the ADMM convergence guarantee~\cite{boyd2011distributed}, where $\mathbf{Z}^{(c)} = \boldsymbol{\Theta}_{\text{com}} + \boldsymbol{S}^{(c)}$ holds in the limit, and provides practical robustness against numerical instability.

\subsection{Multi-task Heads: From Graph Structures to Downstream Predictions}
\label{sec:downstream}

The learned $\boldsymbol{\Theta}_{\text{com}}$ and $\{\boldsymbol{S}^{(c)}\}$ serve as a unified backbone for classification and segmentation. Following~\cite{wu2025ja}, we design a classification head $\mathcal{H}_C$ and a segmentation head $\mathcal{H}_S$ that share the same graph structures but employ distinct inference pathways.

\subsubsection{Classification Head $\mathcal{H}_C$: Generative Discrimination}
\label{sec:cls_head}

For each class $c$, the holistic precision matrix is $\boldsymbol{\Theta}^{(c)} = \boldsymbol{\Theta}_{\text{com}} + \boldsymbol{S}^{(c)}$. Given a test sample's graph node observation $\tilde{\mathbf{z}} \in \mathbb{R}^p$ (post nonparanormal transformation), we compute the log-likelihood score for MAP estimation:
\begin{equation}
\label{eq:cls_head}
s(c \mid \tilde{\mathbf{z}}) = \frac{1}{2}\log\det \boldsymbol{\Theta}^{(c)} - \frac{1}{2}(\tilde{\mathbf{z}} - \hat{\boldsymbol{\mu}}^{(c)})^\top \boldsymbol{\Theta}^{(c)} (\tilde{\mathbf{z}} - \hat{\boldsymbol{\mu}}^{(c)})
\end{equation}
where $\hat{\boldsymbol{\mu}}^{(c)}$ is the class-wise empirical mean computed from the transformed features. The predicted label is $\hat{y} = \arg\max_c\, s(c \mid \tilde{\mathbf{z}})$. This mechanism directly leverages learned graph structures without requiring additional trainable parameters, similar to Gaussian graphical model-based discrimination~\cite{friedman2008sparse}.

\subsubsection{Segmentation Head $\mathcal{H}_S$: Graph-Structured Attention Decoding}
\label{sec:seg_head}

Segmentation requires per-pixel predictions. We exploit the cross-attention matrix $\mathbf{A} \in \mathbb{R}^{p \times N_p}$ as a bidirectional bridge for three-stage decoding, extending the scene labeling framework using sparse precision matrices~\cite{souly2016scene} to the multimodal setting.

\textbf{Stage 1: Graph Message Passing.} Using $\boldsymbol{\Theta}^{(c)}$ as the adjacency structure, we perform message passing that explicitly preserves the sign semantics of conditional dependencies. For each node $i$, we separately aggregate messages based on the topological pathways defined by $\boldsymbol{\Theta}^{(c)}$. Specifically, we route messages from competitive neighbors ($\theta_{ij}^{(c)} > 0$, capturing mutually exclusive roles) and synergistic neighbors ($\theta_{ij}^{(c)} < 0$, capturing semantic co-occurrence):
\begin{equation}
\label{eq:graph_mp_signed}
\mathbf{h}_i = \sigma\!\left( \sum_{j: \theta_{ij}^{(c)} > 0} \alpha_{ij}^{+} \mathbf{W}_{\text{pos}} \tilde{\mathbf{z}}_j + \sum_{j: \theta_{ij}^{(c)} < 0} \alpha_{ij}^{-} \mathbf{W}_{\text{neg}} \tilde{\mathbf{z}}_j \right)
\end{equation}
where the normalized attention weights are defined as:
\begin{equation}
\alpha_{ij}^{+} = \frac{|\theta_{ij}^{(c)}|}{\sum_{k: \theta_{ik}^{(c)} > 0} |\theta_{ik}^{(c)}| + \epsilon}, \quad
\alpha_{ij}^{-} = \frac{|\theta_{ij}^{(c)}|}{\sum_{k: \theta_{ik}^{(c)} < 0} |\theta_{ik}^{(c)}| + \epsilon}
\end{equation}
Here, $\mathbf{W}_{\text{pos}}, \mathbf{W}_{\text{neg}} \in \mathbb{R}^{d_k \times 1}$ are learnable projection weights that independently process the positive ($\theta > 0$) and negative ($\theta < 0$) precision matrix pathways, $\sigma$ is GELU, and $\epsilon$ prevents division by zero. This design ensures that the sign information inherent in the precision matrix—indicating whether two features are mutually exclusive (positive $\theta$) or co-occur (negative $\theta$)—is explicitly preserved throughout the message passing process, rather than being lost via absolute value operations.

\textbf{Stage 2: Node-to-Patch Decoding.} The transposed cross-attention matrix back-projects node-level features to the patch space:
\begin{equation}
\hat{\mathbf{X}}_{\text{seg}} = \mathbf{A}^\top \mathbf{H} \in \mathbb{R}^{N_p \times d_k}, \quad \mathbf{H} = [\mathbf{h}_1; \ldots; \mathbf{h}_p]
\end{equation}

\textbf{Stage 3: Pixel-Level Classification.} A linear layer maps to class probabilities:
\begin{equation}
\hat{\mathbf{Y}} = \text{Softmax}(\hat{\mathbf{X}}_{\text{seg}} \mathbf{W}_s) \in \mathbb{R}^{N_p \times C}
\end{equation}
followed by bilinear upsampling. Training uses pixel-wise cross-entropy loss.

Both heads share the same $\boldsymbol{\Theta}_{\text{com}}$ and $\boldsymbol{S}^{(c)}$ from Sec.~\ref{sec:joint_opt}, with $\mathcal{H}_C$ requiring no additional parameters and $\mathcal{H}_S$ adding only $\mathbf{W}_{\text{pos}}$, $\mathbf{W}_{\text{neg}}$, and $\mathbf{W}_s$.

\subsection{Decoupled Proxy Supervision Strategy}
\label{sec:training_strategy}

Backpropagating downstream task losses through the iterative ADMM solver entails repeated exact matrix eigendecompositions (Eq.~\eqref{eq:z_update}), which is computationally prohibitive and prone to gradient explosion~\cite{boyd2011distributed}. To resolve this, we formulate a \textit{decoupled proxy supervision} strategy, rigorously isolating neural parameter learning ($\mathbf{Q}_{\text{proto}}$, $\mathbf{W}_{\text{pos}}$, $\mathbf{W}_{\text{neg}}$, $\mathbf{W}_s$) from convex graph optimization in three phases:

\textbf{Phase 1: Proxy Supervision.} We optimize the neural parameters directly via a standard pixel-wise cross-entropy proxy task, bypassing the ADMM solver to ensure stable, gradient-driven convergence of the semantic probes.

\textbf{Phase 2: Offline Graph Estimation.} Freezing the network weights (`detach()`), we extract the observation vectors $\mathbf{z}_n$ across the dataset. The empirical covariances $\hat{\boldsymbol{\Sigma}}^{(c)}$ and cross-modal priors $\tilde{\mathbf{W}}^{(c)}$ are computed statically, and the ADMM algorithm (Algorithm~\ref{alg:framework}) is executed offline to global convergence.

\textbf{Phase 3: Graph-Guided Inference.} The learned static topologies ($\boldsymbol{\Theta}^{(c)} = \boldsymbol{\Theta}_{\text{com}} + \boldsymbol{S}^{(c)}$) are explicitly injected back into the multi-task heads as fixed priors to govern generative classification (Eq.~\eqref{eq:cls_head}) and topology-aware message passing (Eq.~\eqref{eq:graph_mp_signed}).

\textbf{Remark on Suboptimality.} While this decoupled paradigm cleanly circumvents unrolled optimization instabilities, it theoretically sacrifices a strictly end-to-end global optimum. However, this suboptimality gap is mathematically mitigated by our nonparanormal transformation (Sec.~\ref{sec:nonparanormal}). By bounding the empirical distributions into a standardized Gaussian space, we significantly suppress representation drift. This ensures that proxy-learned features robustly support the offline Markovian topology estimation, gracefully trading marginal global optimality for guaranteed convergence and numerical stability.
\section{Experiments}
\label{sec:experiments}

\subsection{Experimental Setup}

\textbf{Datasets:} We evaluate on eight benchmarks: CIFAR-10/100~\cite{krizhevsky2009learning}, CUB-200-2011~\cite{wah2011cub}, and Caltech-256~\cite{griffin2007caltech256} for classification; PASCAL VOC 2012~\cite{everingham2012pascal}, ADE20K~\cite{zhou2019semantic}, MS COCO 2014~\cite{lin2014microsoft}, and Kvasir-SEG~\cite{jha2020kvasir} for segmentation. For vision-only datasets, class-attribute texts are generated via Qwen3-VL. Details are in the supplement.

\textbf{Implementation Details:} The unified encoder is SigLIP 2 ViT-B/16~\cite{tschannen2025siglip2} ($d=768$, $N_p=196$) with $224 \times 224$ inputs. Text is rendered via PIL. We strictly enforce $p < n_c$ to resolve the high-dimensional low-sample-size (HDLSS) bottleneck. Joint optimization uses candidate $k \in \{0..50\}$, $\mu=1.0$, $\gamma=0.5$, and max 200 ADMM iterations. Hyperparameters $\rho, \gamma_s \in \{0.01, 0.05, 0.1, 0.2\}$ are grid-searched. Experiments run on an NVIDIA A800 GPU. We compare against task-specific SOTA architectures and multimodal paradigms.

\subsection{Main Results}
\label{sec:main_results}

\textbf{Classification:} Table~\ref{tab:cls_results} shows CM-GLasso consistently achieves SOTA performance. On fine-grained CUB-200-2011, it attains \textbf{92.83\%} accuracy, outperforming PRO-VPT~\cite{shang2025provpt} by 1.13\%, proving the estimated sparse semantic topology provides a stronger structural inductive bias than pure prompt tuning. It also leads on CIFAR-10 (\textbf{94.71\%}), CIFAR-100 (\textbf{94.26\%}), and Caltech-256 (\textbf{86.07\%}), with robust F1 scores under class imbalances.

\begin{table}[htbp]
\centering
\caption{Comparative results on classification tasks.}
\label{tab:cls_results}
\small
\setlength{\tabcolsep}{8pt}
\begin{tabular}{@{}llcc@{}}
\toprule
\textbf{Dataset} & \textbf{Method} & \textbf{F1} & \textbf{ACC} \\
\midrule
\multirow{6}{*}{\textbf{CUB-200-2011}} 
& ShuffleNetV2~\cite{ma2018shufflenetv2} & 0.8774 & 0.8763 \\
& DA-VPT~\cite{ren2025davpt} & --- & 0.9130 \\
& PRO-VPT~\cite{shang2025provpt} & --- & 0.9170 \\
& VFPT~\cite{zeng2024vfpt} & --- & 0.9050 \\
& MT-ASM~\cite{liu2024mtasm} & --- & 0.8800 \\
& \textbf{CM-GLasso (Ours)} & \textbf{0.8836} & \textbf{0.9283} \\
\midrule
\multirow{5}{*}{\textbf{CIFAR-10}} 
& PCA-CNN-DenseNet~\cite{alharis2024pca} & --- & 0.8982 \\
& Nddr-cnn~\cite{gao2019nddr} & --- & 0.8853 \\
& self-defined MTL~\cite{hyun2024selfdefined} & --- & 0.8494 \\
& OnPro-0.5k~\cite{wei2023onpro} & --- & 0.7260 \\
& \textbf{CM-GLasso (Ours)} & \textbf{0.9309} & \textbf{0.9471} \\
\midrule
\multirow{5}{*}{\textbf{CIFAR-100}} 
& PALM~\cite{lu2024palm} & --- & 0.7820 \\
& SSF~\cite{lian2022ssf} & --- & 0.9399 \\
& Astroformer~\cite{dagli2023astroformer} & --- & 0.9360 \\
& SPT-Swin~\cite{ferdous2024sptswin} & 0.9295 & 0.9295 \\
& \textbf{CM-GLasso (Ours)} & \textbf{0.9300} & \textbf{0.9426} \\
\midrule
\multirow{4}{*}{\textbf{Caltech-256}} 
& TMC~\cite{liu2023tmc} & --- & 0.8364 \\
& CPC~\cite{zhi2024cpc} & --- & 0.8550 \\
& EEG-VGG Fusion~\cite{jahanaray2025eeg} & --- & 0.8100 \\
& \textbf{CM-GLasso (Ours)} & \textbf{0.8528} & \textbf{0.8607} \\
\bottomrule
\end{tabular}
\end{table}

\textbf{Semantic Segmentation:} Table~\ref{tab:seg_results} confirms the precision matrix $\boldsymbol{\Theta}^{(c)}$ furnishes a highly robust topology for pixel-level prediction. CM-GLasso achieves \textbf{64.01\%} mIoU on ADE20K (surpassing InternImage-H~\cite{wang2023internimage}), \textbf{74.75\%} on VOC-2012, and \textbf{46.82\%} on COCO-2014. In medical imaging, it attains \textbf{89.03\%} on Kvasir-SEG, outperforming PolypMixNet~\cite{jia2024polypmixnet} and validating cross-domain adaptability. We further discuss integrating a U-Net decoder into $\mathcal{H}_S$ in the supplement.

\begin{table}[htbp]
\centering
\caption{Comparative results on semantic segmentation (mIoU).}
\label{tab:seg_results}
\small
\setlength{\tabcolsep}{12pt}
\begin{tabular}{@{}llc@{}}
\toprule
\textbf{Dataset} & \textbf{Method} & \textbf{mIoU} \\
\midrule
\multirow{4}{*}{\textbf{ADE20K}} 
& OneFormer~\cite{jain2023oneformer} & 0.5700 \\
& InternImage-H~\cite{wang2023internimage} & 0.6290 \\
& OmniVec2~\cite{srivastava2024omnivec2} & 0.5850 \\
& \textbf{CM-GLasso (Ours)} & \textbf{0.6401} \\
\midrule
\multirow{4}{*}{\textbf{Kvasir-SEG}} 
& Polyp-PVT~\cite{dong2023polypvt} & 0.8640 \\
& PolypMixNet~\cite{jia2024polypmixnet} & 0.8885 \\
& MedFoundX~\cite{shawon2025medfoundx} & 0.8668 \\
& \textbf{CM-GLasso (Ours)} & \textbf{0.8903} \\
\midrule
\multirow{4}{*}{\textbf{VOC-2012}} 
& AuxSegNet+~\cite{xu2024auxsegnet} & 0.7090 \\
& GroupViT~\cite{xu2022groupvit} & 0.5230 \\
& PrivObNet~\cite{tay2024privobfnet} & 0.7150 \\
& \textbf{CM-GLasso (Ours)} & \textbf{0.7475} \\
\midrule
\multirow{4}{*}{\textbf{COCO-2014}} 
& MulP-VSS~\cite{duan2025mulpvss} & 0.4660 \\
& CLIP-ES~\cite{lin2023clipes} & 0.4540 \\
& BECO~\cite{rong2023beco} & 0.4510 \\
& \textbf{CM-GLasso (Ours)} & \textbf{0.4682} \\
\bottomrule
\end{tabular}
\end{table}

\subsection{Ablation Studies}
\label{sec:ablation}

To comprehensively validate our framework, Tables~\ref{tab:abla_text}--\ref{tab:abla_ebic} present key ablations (arithmetic means across all datasets) in sequential order.

\begin{table}[htbp]
\centering
\caption{Ablation: Text Encoding Strategy.}
\label{tab:abla_text}
\small
\setlength{\tabcolsep}{8pt}
\begin{tabular}{@{}lcc@{}}
\toprule
\textbf{Encoder} & \textbf{ACC} & \textbf{mIoU} \\
\midrule
BERT+ViT (Het.) & 84.23 & 53.27 \\
CLIP text+ViT~\cite{radford2021learning} & 88.02 & 59.89 \\
\cellcolor{gray!12}\textbf{Render+SigLIP 2} & \cellcolor{gray!12}\textbf{91.97} & \cellcolor{gray!12}\textbf{68.65} \\
\bottomrule
\end{tabular}
\end{table}

\vspace{5mm}  

\begin{table}[htbp]
\centering
\caption{Ablation: Patch-to-Node Mapping.}
\label{tab:abla_mapping}
\small
\setlength{\tabcolsep}{8pt}
\begin{tabular}{@{}lcc@{}}
\toprule
\textbf{Strategy} & \textbf{ACC} & \textbf{mIoU} \\
\midrule
PCA ($768\!\to\!p$) & 70.86 & 44.18 \\
Linear FC & 87.37 & 62.93 \\
\cellcolor{gray!12}\textbf{Cross-Attn (Ours)} & \cellcolor{gray!12}\textbf{91.97} & \cellcolor{gray!12}\textbf{68.65} \\
\bottomrule
\end{tabular}
\end{table}

\vspace{5mm}  

\begin{table}[htbp]
\centering
\caption{Ablation: Nonparanormal Transform.}
\label{tab:abla_npn}
\small
\setlength{\tabcolsep}{8pt}
\begin{tabular}{@{}lccc@{}}
\toprule
\textbf{Status} & \textbf{SW Pass} & \textbf{ACC} & \textbf{mIoU} \\
\midrule
w/o Trans. & $\sim$23\% & 83.58 & 59.02 \\
\cellcolor{gray!12}\textbf{w/ Trans.} & \cellcolor{gray!12}\textbf{$\sim$88\%} & \cellcolor{gray!12}\textbf{91.97} & \cellcolor{gray!12}\textbf{68.65} \\
\bottomrule
\end{tabular}
\end{table}

\vspace{5mm}  

\begin{table}[htbp]
\centering
\caption{Ablation: Optimization (CSR: Common Ratio).}
\label{tab:abla_opt}
\small
\setlength{\tabcolsep}{8pt}
\begin{tabular}{@{}lcc@{}}
\toprule
\textbf{Method} & \textbf{CSR} & \textbf{Gen. Gap} \\
\midrule
Indep. GLasso~\cite{friedman2008sparse} & --- & 8.29\% \\
Two-stage~\cite{danaher2014joint} & 0.37 & 3.04\% \\
\cellcolor{gray!12}\textbf{Joint ADMM} & \cellcolor{gray!12}\textbf{0.42} & \cellcolor{gray!12}\textbf{1.93\%} \\
\bottomrule
\end{tabular}
\end{table}

\vspace{5mm}  

\begin{table}[htbp]
\centering
\caption{Ablation: Task Head Precision Matrix.}
\label{tab:abla_matrix}
\small
\setlength{\tabcolsep}{8pt}
\begin{tabular}{@{}lcc@{}}
\toprule
\textbf{Matrix Used} & \textbf{ACC} & \textbf{mIoU} \\
\midrule
Only $\boldsymbol{\Theta}_{\text{com}}$ & 84.82 & 63.17 \\
Only $\boldsymbol{S}^{(c)}$ & 88.43 & 65.58 \\
\cellcolor{gray!12}$\boldsymbol{\Theta}_{\text{com}} + \boldsymbol{S}^{(c)}$ & \cellcolor{gray!12}\textbf{91.97} & \cellcolor{gray!12}\textbf{68.65} \\
\bottomrule
\end{tabular}
\end{table}

\vspace{5mm}  

\begin{table}[htbp]
\centering
\caption{Ablation: Prior Selection via eBIC.}
\label{tab:abla_ebic}
\small
\setlength{\tabcolsep}{8pt}
\begin{tabular}{@{}lccc@{}}
\toprule
\textbf{Direction} & $\bar{k}^*$ & \multicolumn{2}{c}{$\boldsymbol{k^*\!=\!0}$ \textbf{Ratio}} \\
\midrule
Text $\to$ Image & 15.6 & \multicolumn{2}{c}{13.8\%} \\
Image $\to$ Text & 6.7 & \multicolumn{2}{c}{33.1\%} \\
\cellcolor{gray!12}\textbf{Self-Priors} & \cellcolor{gray!12}\textbf{$\approx$0.1} & \multicolumn{2}{c}{\cellcolor{gray!12}\textbf{>84.0\%}} \\
\bottomrule
\end{tabular}
\end{table}

\vspace{5mm}  

\begin{table}[htbp]
\centering
\caption{Detailed metrics for Patch-to-Node mapping strategies. $|E|$ denotes the average number of edges.}
\label{tab:spca_detail}
\small
\begin{tabular}{@{}lccc@{}}
\toprule
\textbf{Metric} & \textbf{PCA} & \textbf{FC} & \textbf{Ours} \\
\midrule
\multicolumn{4}{@{}l}{\textit{Graph Structure Quality}} \\
\quad Avg. Edges $|E|$ in $\boldsymbol{\Theta}^{(c)}$ & 987 & 491 & \textbf{238} \\
\quad Spurious Edge Ratio (\%) $\downarrow$ & 68.7 & 31.2 & \textbf{11.4} \\
\quad $\bar{k}^*$ (Prior Utilization) & 1.2 & 5.8 & \textbf{15.6} \\
\midrule
\multicolumn{4}{@{}l}{\textit{Interpretability \& Architecture}} \\
\quad Native $p \times p$ Prior & $\times$ & $\times$ & $\checkmark$ \\
\quad Supports $\mathbf{A}^\top$ Back-proj. & $\times$ & $\times$ & $\checkmark$ \\
\bottomrule
\end{tabular}
\end{table}

\textbf{Feature Extraction \& Mapping:} The Render+SigLIP 2 paradigm (Table~\ref{tab:abla_text}) maintains the most compact parameter footprint (84M vs. 196M for Hetero.) while elevating prior reliability ($\bar{k}^*=15.6$), drastically improving $\mathcal{H}_S$. Compared to PCA/FC (Table~\ref{tab:abla_mapping}), our cross-attention distillation uniquely yields a native $p \times p$ prior alignment and supports $\mathbf{A}^\top$ back-projection without auxiliary layers. As detailed in Table~\ref{tab:spca_detail}, it crucially generates the sparsest graph structure (Avg. $|E|=238$ vs. 987) and the lowest spurious edge ratio (11.4\% vs. 68.7\%), faithfully characterizing true conditional dependencies and thereby avoiding the propagation of noisy topological signals.

\textbf{Statistical \& Optimization Properties:} The nonparanormal transformation directly improves estimation quality by raising the Shapiro-Wilk Gaussianity pass rate to $\sim$88\% (Table~\ref{tab:abla_npn}). Furthermore, Joint ADMM limits the generalization gap to 1.93\% (Table~\ref{tab:abla_opt}). This is because joint optimization allocates 42\% of edges to $\boldsymbol{\Theta}_{\text{com}}$, serving as a robust cross-class regularizer that prevents overfitting in low-sample categories.

\begin{table}[htbp]
\centering
\caption{Sensitivity analysis of $\rho$ and $\gamma_s$. Values are presented as ACC(\%) / mIoU(\%).}
\label{tab:hyperparam}
\small
\begin{tabular}{@{}lcccc@{}}
\toprule
$\rho \downarrow$ \textbackslash ~$\gamma_s \rightarrow$ & 0.01 & 0.05 & 0.10 & 0.20 \\
\midrule
0.01 & 90.53/66.94 & 91.02/67.53 & 91.18/67.72 & 90.61/67.03 \\
0.05 & 91.27/67.68 & 91.64/68.27 & \textbf{91.97/68.65} & 91.41/67.91 \\
0.10 & 91.12/67.56 & 91.53/68.14 & 91.82/68.47 & 91.23/67.63 \\
0.20 & 90.46/66.81 & 90.83/67.24 & 91.04/67.40 & 90.21/66.59 \\
\bottomrule
\end{tabular}
\end{table}

\textbf{Task Design \& eBIC Tautology Prevention:} Both task heads peak when utilizing the combined matrix $\boldsymbol{\Theta}^{(c)}$ (Table~\ref{tab:abla_matrix}), perfectly balancing shared foundational structures with class-specific discriminability. Table~\ref{tab:abla_ebic} demonstrates eBIC's rigorous prevention of circular reasoning: while text priors heavily guide images ($\bar{k}^*=15.6$), intra-modal self-priors are universally rejected ($k^* \approx 0$ for $>84\%$ of cases), gracefully degrading to standard unbiased optimization. Finally, as shown in Table~\ref{tab:hyperparam}, the exhaustive grid search over $\rho$ and $\gamma_s$ demonstrates that within the optimal range of $[0.05, 0.10]$, performance fluctuations remain under 1\%, proving that the synergy between ADMM and the adaptive eBIC mechanism significantly minimizes the need for exhaustive manual tuning.

\subsection{Visualization and Complexity}
\label{sec:vis_complexity}

\begin{figure*}[t!]
  \centering

  \newlength{\clsw}
  \setlength{\clsw}{0.108\textwidth}

  \setlength{\tabcolsep}{1.2pt}
  \renewcommand{\arraystretch}{0.6}

  \begin{tabular}{
    @{}
    r
    c@{\hskip 1.2pt}c@{\hskip 1.2pt}c@{\hskip 1.2pt}c
    @{\hskip 5pt}
    c@{\hskip 1.2pt}c@{\hskip 1.2pt}c@{\hskip 1.2pt}c
    @{}}

    &
    \multicolumn{4}{c}{\scriptsize\textbf{Input Images}} &
    \multicolumn{4}{c}{\scriptsize\textbf{GAM Visualizations}} \\[3pt]

    \raisebox{0.45\clsw}[0pt][0pt]{\rotatebox[origin=c]{90}{\scriptsize\textbf{CUB-200}}} &
    \includegraphics[width=\clsw,height=\clsw]{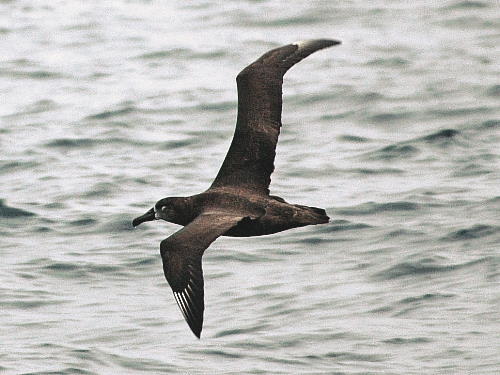} &
    \includegraphics[width=\clsw,height=\clsw]{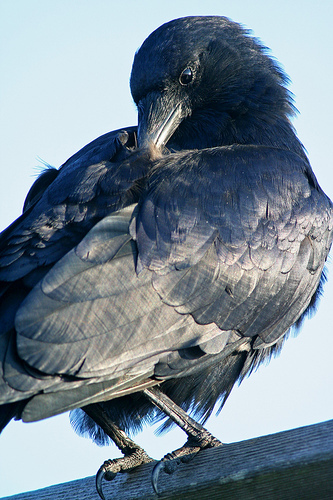} &
    \includegraphics[width=\clsw,height=\clsw]{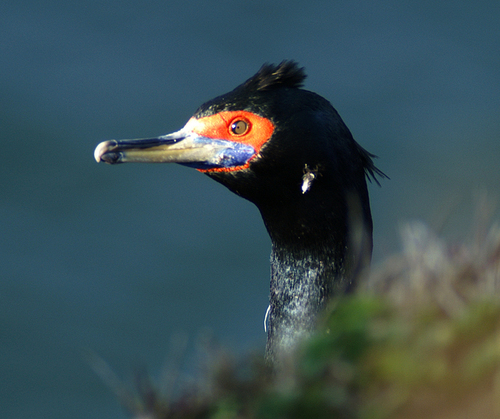} &
    \includegraphics[width=\clsw,height=\clsw]{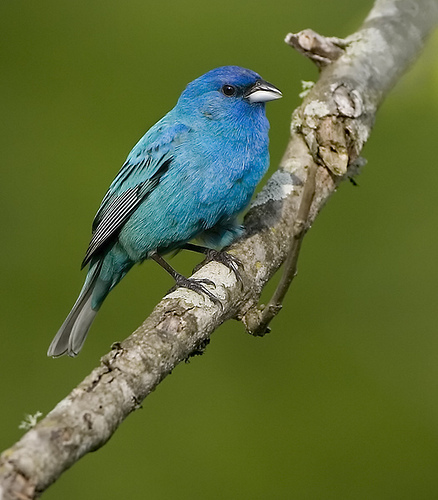} &
    \includegraphics[width=\clsw,height=\clsw]{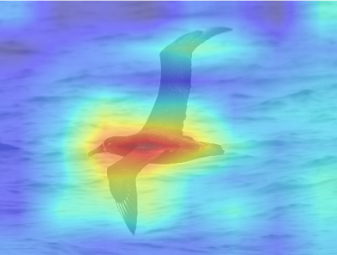} &
    \includegraphics[width=\clsw,height=\clsw]{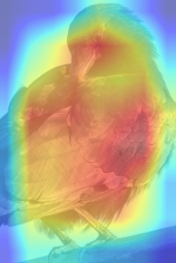} &
    \includegraphics[width=\clsw,height=\clsw]{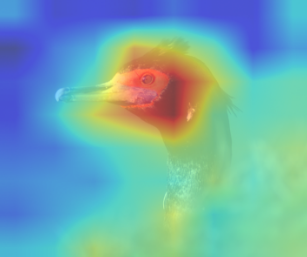} &
    \includegraphics[width=\clsw,height=\clsw]{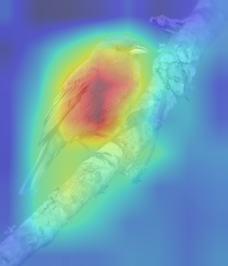} \\[2pt]

    \raisebox{0.45\clsw}[0pt][0pt]{\rotatebox[origin=c]{90}{\scriptsize\textbf{CIFAR-10}}} &
    \includegraphics[width=\clsw,height=\clsw]{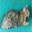} &
    \includegraphics[width=\clsw,height=\clsw]{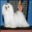} &
    \includegraphics[width=\clsw,height=\clsw]{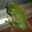} &
    \includegraphics[width=\clsw,height=\clsw]{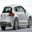} &
    \includegraphics[width=\clsw,height=\clsw]{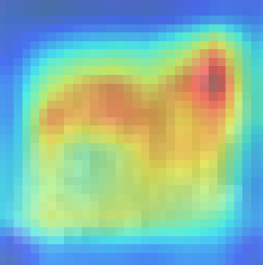} &
    \includegraphics[width=\clsw,height=\clsw]{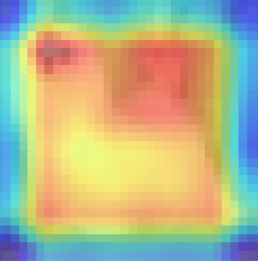} &
    \includegraphics[width=\clsw,height=\clsw]{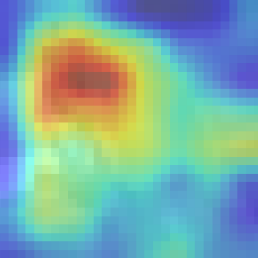} &
    \includegraphics[width=\clsw,height=\clsw]{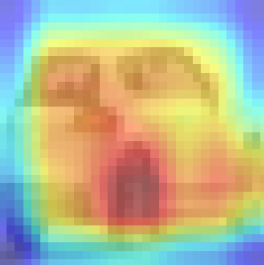} \\[2pt]

    \raisebox{0.45\clsw}[0pt][0pt]{\rotatebox[origin=c]{90}{\scriptsize\textbf{CIFAR-100}}} &
    \includegraphics[width=\clsw,height=\clsw]{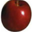} &
    \includegraphics[width=\clsw,height=\clsw]{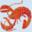} &
    \includegraphics[width=\clsw,height=\clsw]{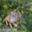} &
    \includegraphics[width=\clsw,height=\clsw]{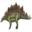} &
    \includegraphics[width=\clsw,height=\clsw]{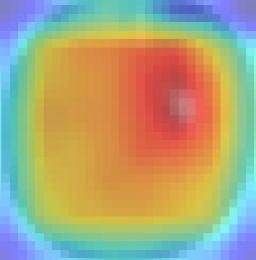} &
    \includegraphics[width=\clsw,height=\clsw]{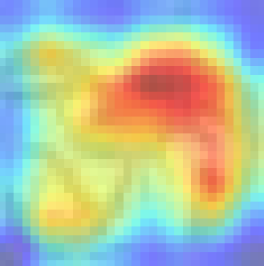} &
    \includegraphics[width=\clsw,height=\clsw]{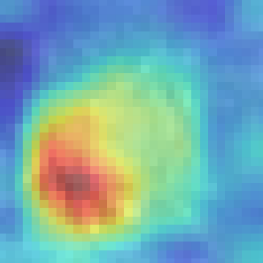} &
    \includegraphics[width=\clsw,height=\clsw]{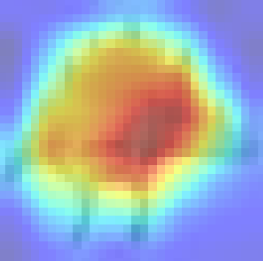} \\[2pt]

    \raisebox{0.45\clsw}[0pt][0pt]{\rotatebox[origin=c]{90}{\scriptsize\textbf{Caltech-256}}} &
    \includegraphics[width=\clsw,height=\clsw]{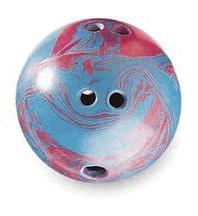} &
    \includegraphics[width=\clsw,height=\clsw]{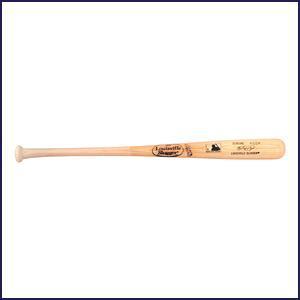} &
    \includegraphics[width=\clsw,height=\clsw]{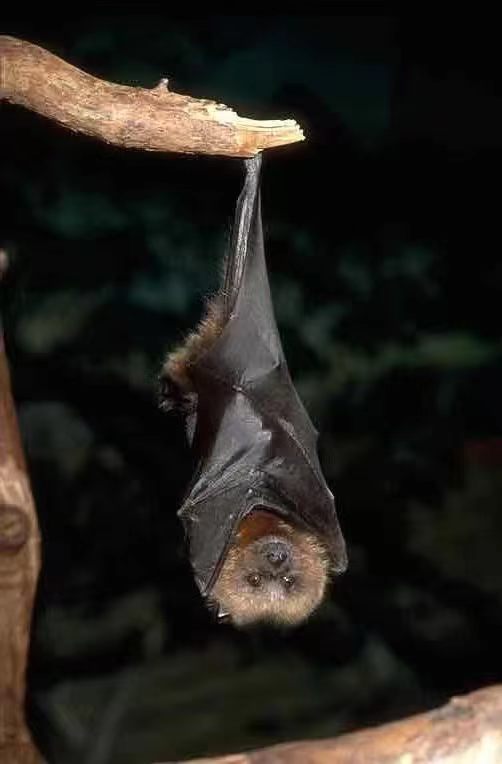} &
    \includegraphics[width=\clsw,height=\clsw]{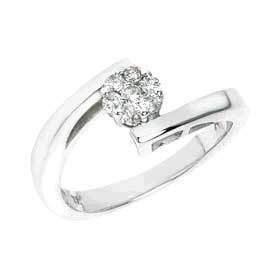} &
    \includegraphics[width=\clsw,height=\clsw]{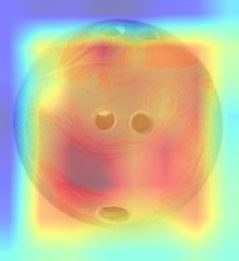} &
    \includegraphics[width=\clsw,height=\clsw]{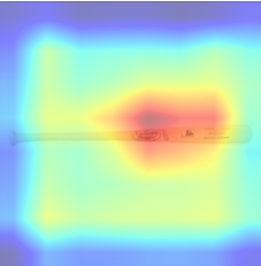} &
    \includegraphics[width=\clsw,height=\clsw]{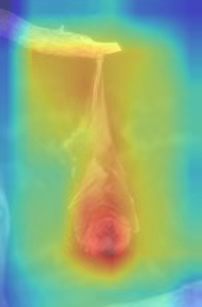} &
    \includegraphics[width=\clsw,height=\clsw]{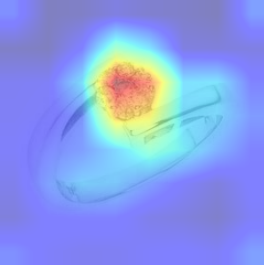} \\

  \end{tabular}

  \vspace{1mm}
  \hfill
  \makebox[0.48\textwidth][c]{%
    $\underbrace{\hspace{0.44\textwidth}}_{\text{\scriptsize Input Images}}$}%
  \hfill
  \makebox[0.48\textwidth][c]{%
    $\underbrace{\hspace{0.44\textwidth}}_{\text{\scriptsize GAM Heatmaps}}$}%
  \hfill\null

  \vspace{1mm}
  \caption{%
    \textbf{GAM visualization of the classification head $\mathcal{H}_C$.}
    For each dataset (CUB-200-2011, CIFAR-10, CIFAR-100, Caltech-256),
    four input images (left block) are paired with their GAM heatmaps (right block).
    Warm regions (red/yellow) indicate spatially discriminative areas;
    cool regions (blue) indicate low contribution.
    CM-GLasso consistently focuses on class-discriminative semantics
    (\emph{e.g.}, bird head/wings, vehicle contours) rather than background noise,
    validating that cross-modal prior guidance improves classification interpretability.%
  }
  \label{fig:gam_vis}
\end{figure*}

\begin{figure*}[t!]
  \centering

  \newlength{\segw}
  \setlength{\segw}{0.095\textwidth}

  \setlength{\tabcolsep}{0.8pt}
  \renewcommand{\arraystretch}{0.5}

  \begin{tabular}{
    @{}
    r
    c@{\hskip 0.8pt}c@{\hskip 0.8pt}c
    @{\hskip 4pt}
    c@{\hskip 0.8pt}c@{\hskip 0.8pt}c
    @{\hskip 4pt}
    c@{\hskip 0.8pt}c@{\hskip 0.8pt}c
    @{}}

    &
    \multicolumn{3}{c}{\scriptsize\textbf{Sample 1}} &
    \multicolumn{3}{c}{\scriptsize\textbf{Sample 2}} &
    \multicolumn{3}{c}{\scriptsize\textbf{Sample 3}} \\[-0.5pt]

    &
    {\tiny Image} & {\tiny GT} & {\tiny Ours} &
    {\tiny Image} & {\tiny GT} & {\tiny Ours} &
    {\tiny Image} & {\tiny GT} & {\tiny Ours} \\[2pt]

    \raisebox{0.45\segw}[0pt][0pt]{\rotatebox[origin=c]{90}{\scriptsize\textbf{ADE20K}}} &
    \includegraphics[width=\segw,height=\segw]{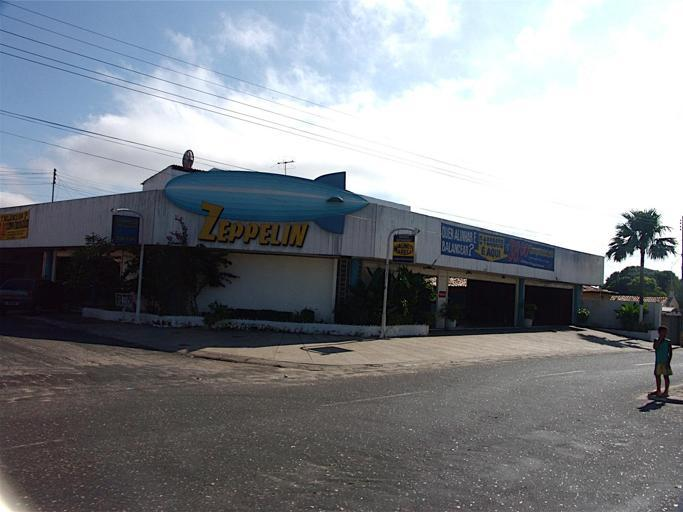} &
    \includegraphics[width=\segw,height=\segw]{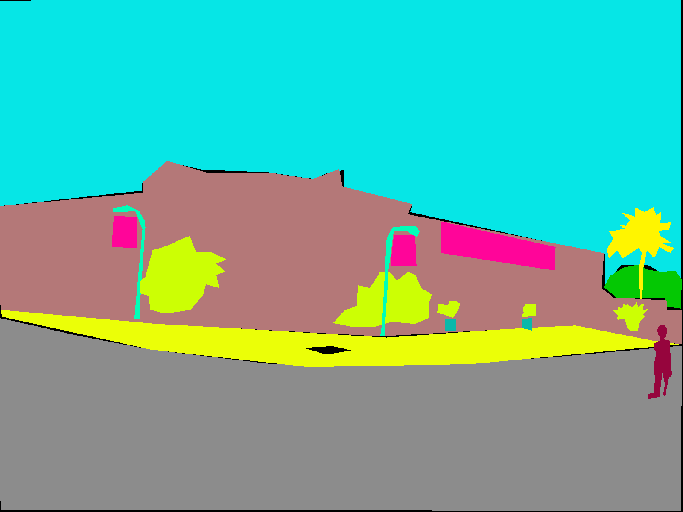} &
    \includegraphics[width=\segw,height=\segw]{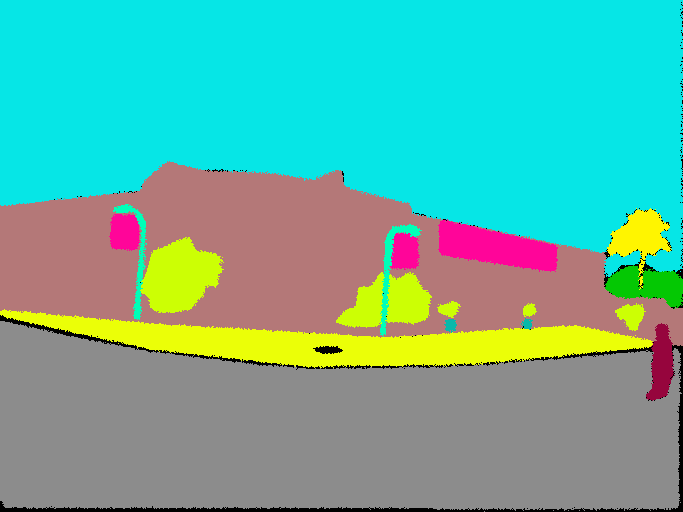} &
    \includegraphics[width=\segw,height=\segw]{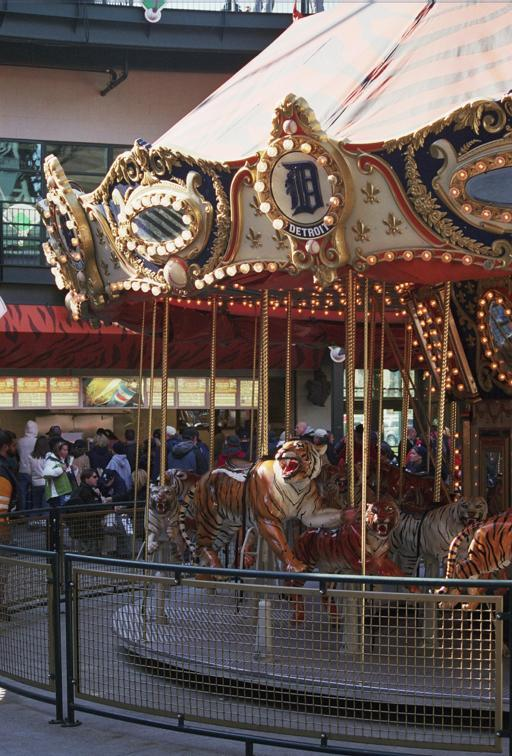} &
    \includegraphics[width=\segw,height=\segw]{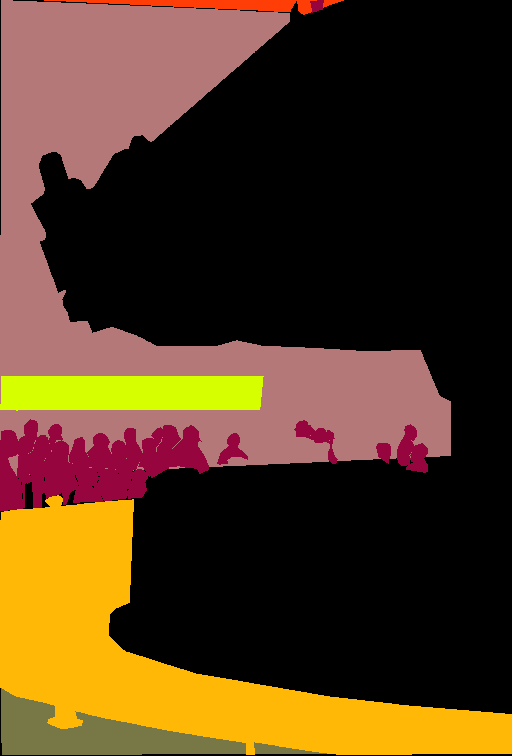} &
    \includegraphics[width=\segw,height=\segw]{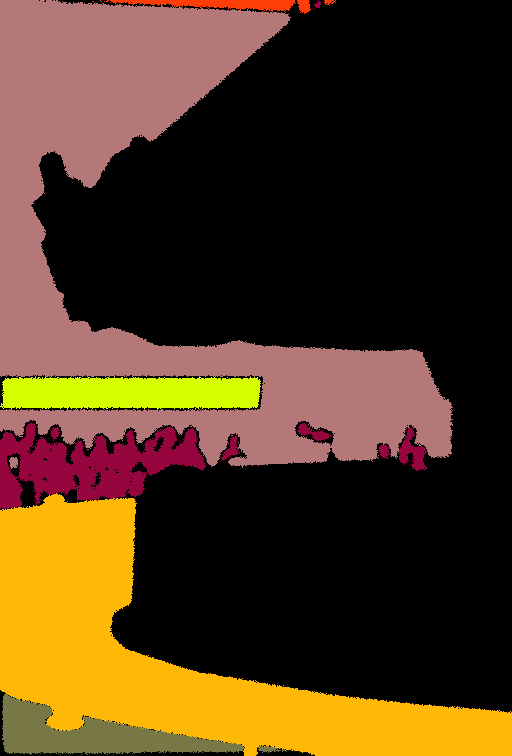} &
    \includegraphics[width=\segw,height=\segw]{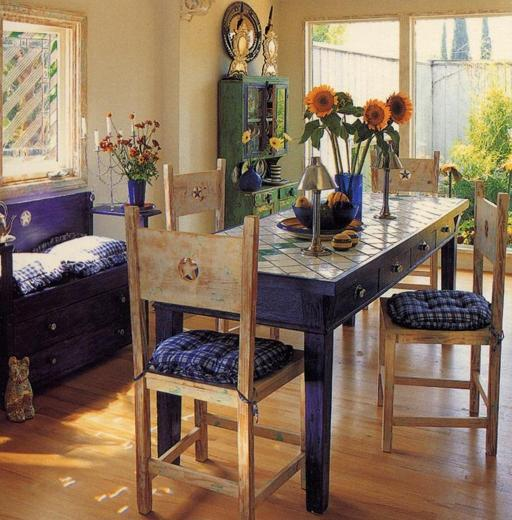} &
    \includegraphics[width=\segw,height=\segw]{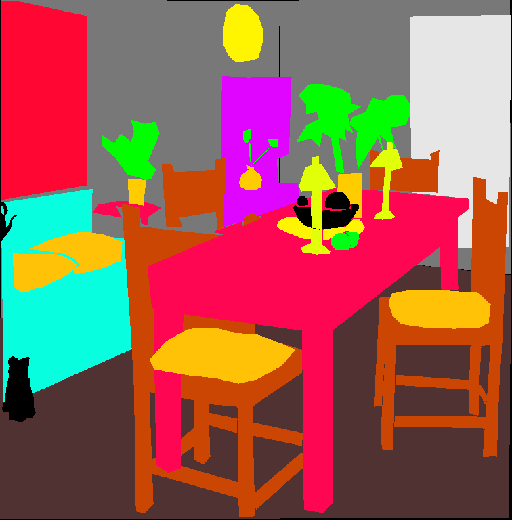} &
    \includegraphics[width=\segw,height=\segw]{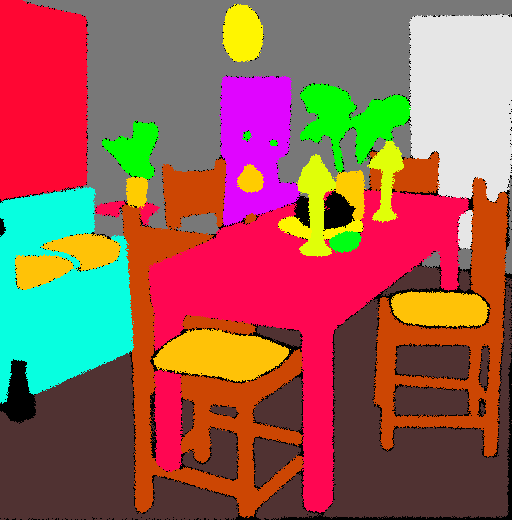} \\[2pt]

    \raisebox{0.45\segw}[0pt][0pt]{\rotatebox[origin=c]{90}{\scriptsize\textbf{Kvasir}}} &
    \includegraphics[width=\segw,height=\segw]{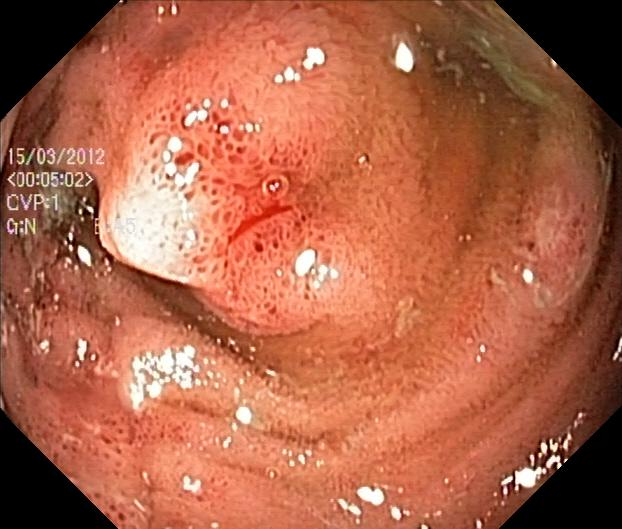} &
    \includegraphics[width=\segw,height=\segw]{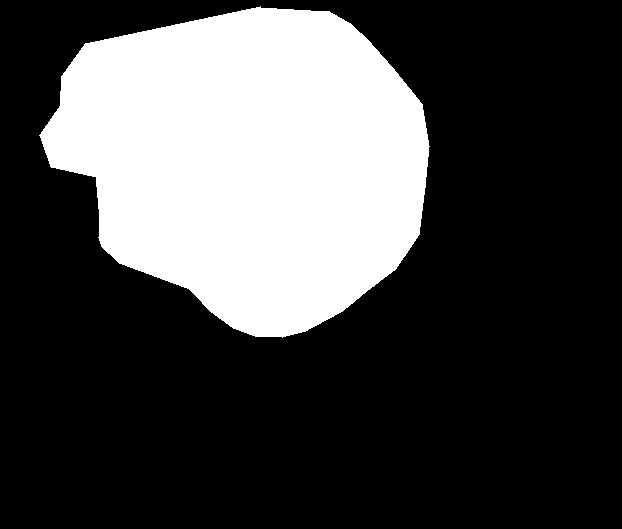} &
    \includegraphics[width=\segw,height=\segw]{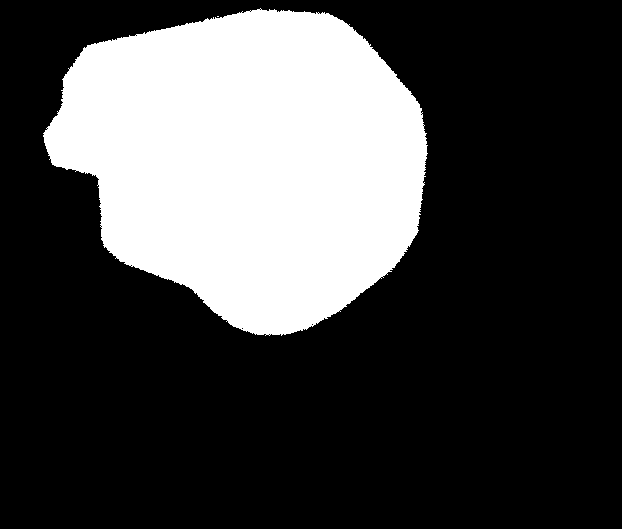} &
    \includegraphics[width=\segw,height=\segw]{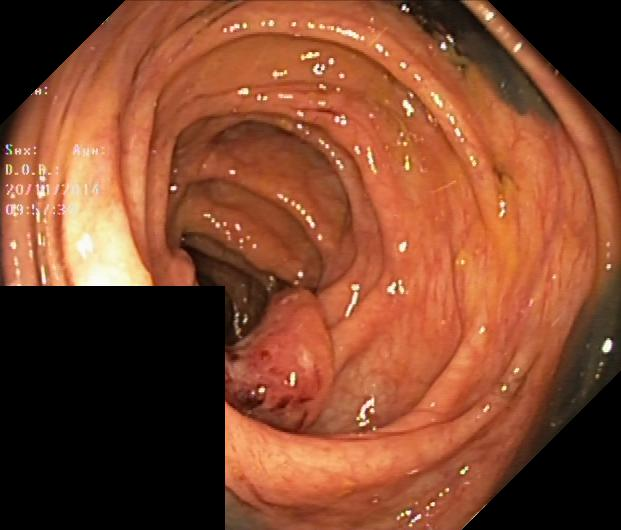} &
    \includegraphics[width=\segw,height=\segw]{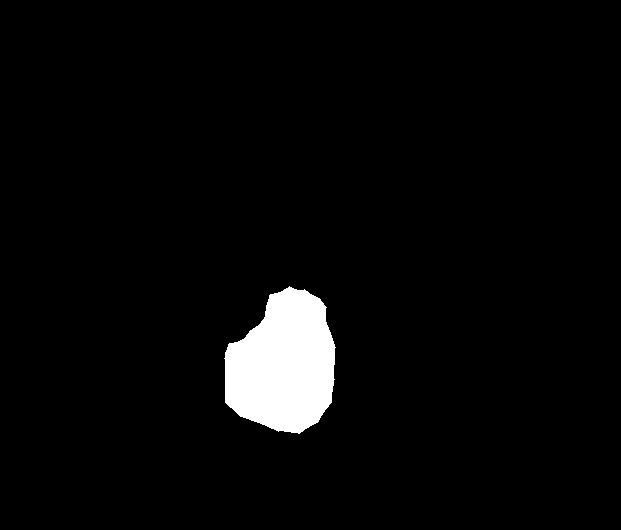} &
    \includegraphics[width=\segw,height=\segw]{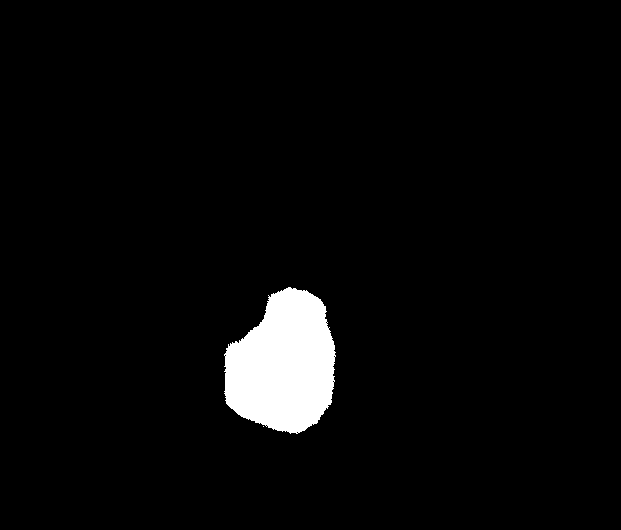} &
    \includegraphics[width=\segw,height=\segw]{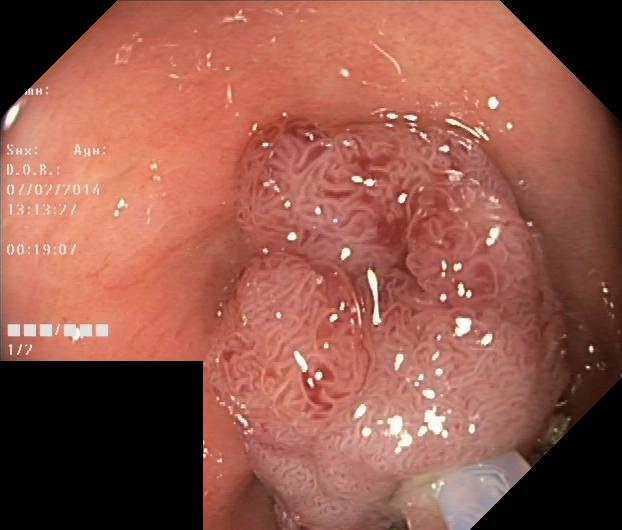} &
    \includegraphics[width=\segw,height=\segw]{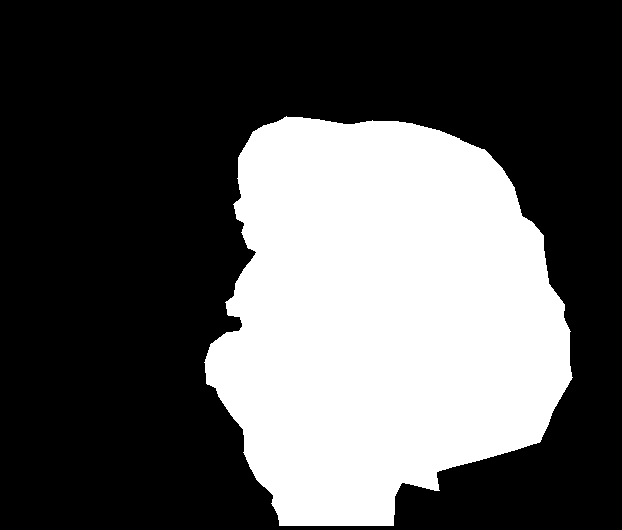} &
    \includegraphics[width=\segw,height=\segw]{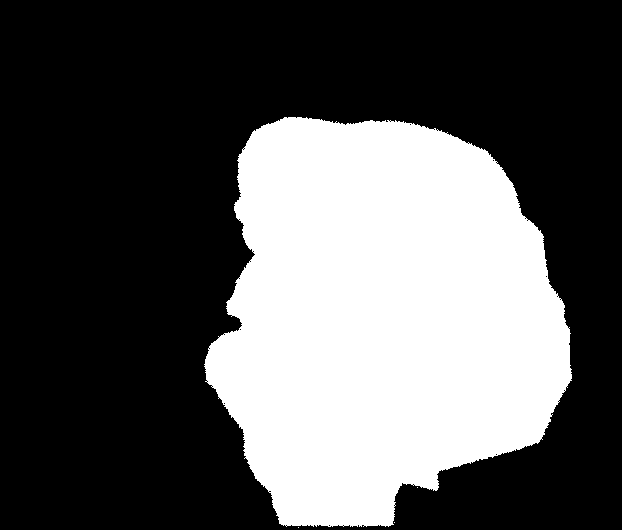} \\[2pt]

    \raisebox{0.45\segw}[0pt][0pt]{\rotatebox[origin=c]{90}{\scriptsize\textbf{VOC-2012}}} &
    \includegraphics[width=\segw,height=\segw]{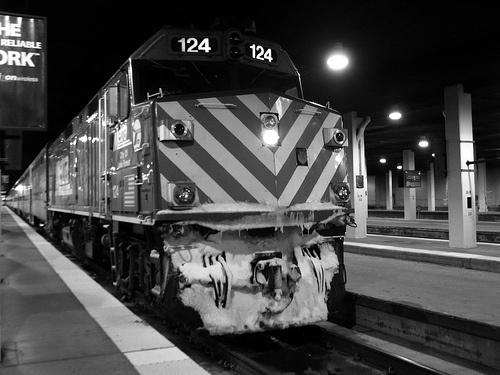} &
    \includegraphics[width=\segw,height=\segw]{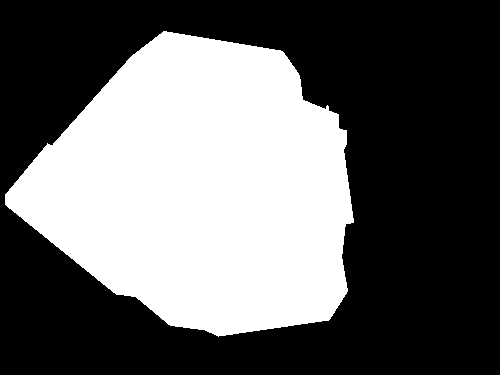} &
    \includegraphics[width=\segw,height=\segw]{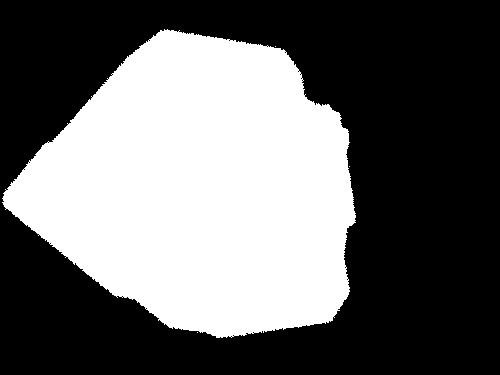} &
    \includegraphics[width=\segw,height=\segw]{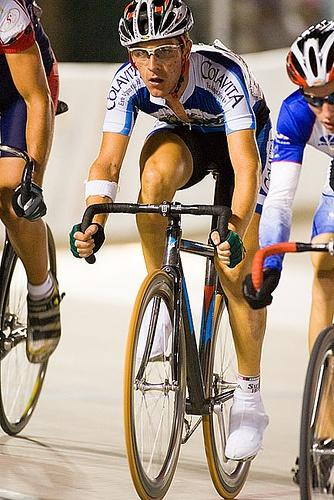} &
    \includegraphics[width=\segw,height=\segw]{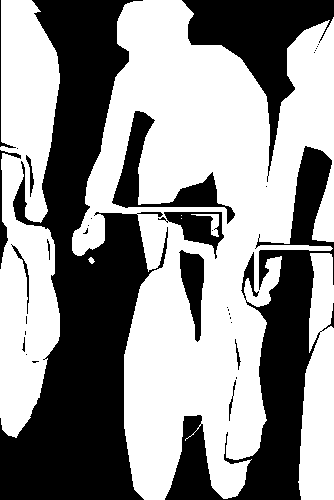} &
    \includegraphics[width=\segw,height=\segw]{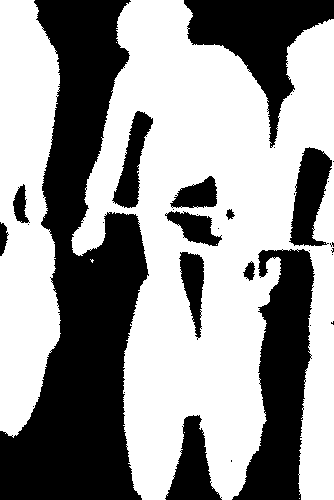} &
    \includegraphics[width=\segw,height=\segw]{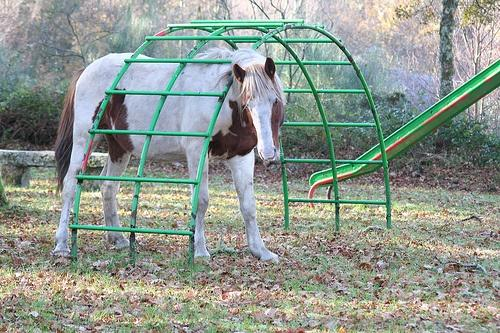} &
    \includegraphics[width=\segw,height=\segw]{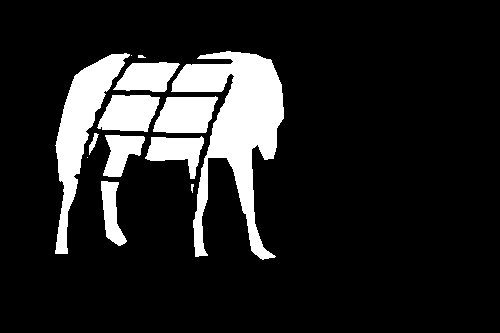} &
    \includegraphics[width=\segw,height=\segw]{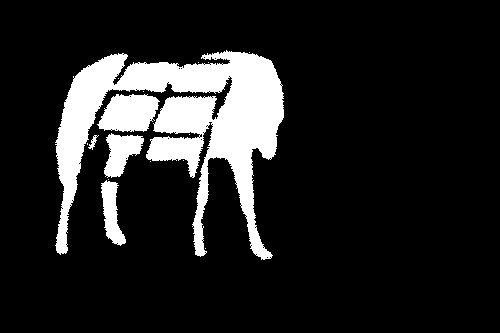} \\[2pt]

    \raisebox{0.45\segw}[0pt][0pt]{\rotatebox[origin=c]{90}{\scriptsize\textbf{COCO}}} &
    \includegraphics[width=\segw,height=\segw]{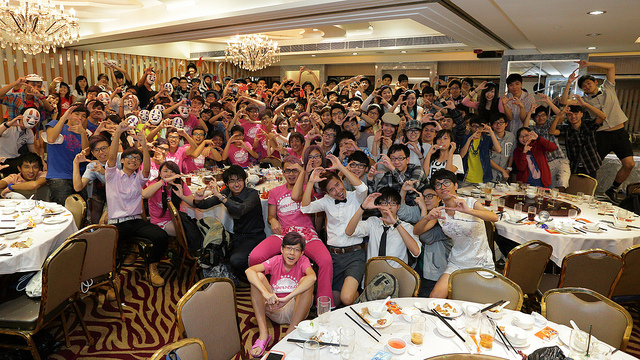} &
    \includegraphics[width=\segw,height=\segw]{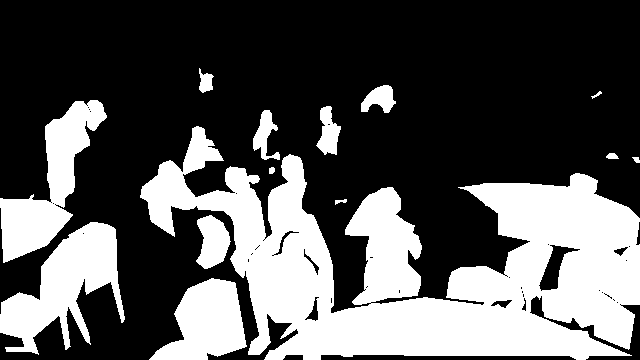} &
    \includegraphics[width=\segw,height=\segw]{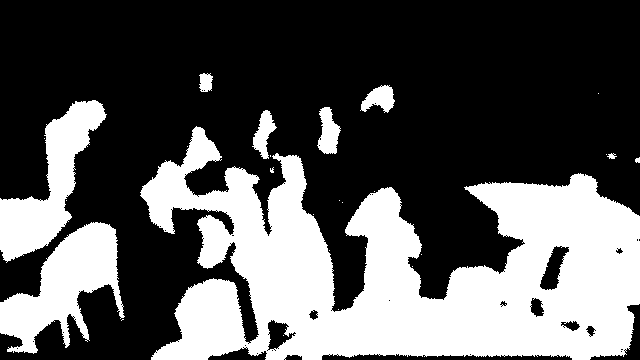} &
    \includegraphics[width=\segw,height=\segw]{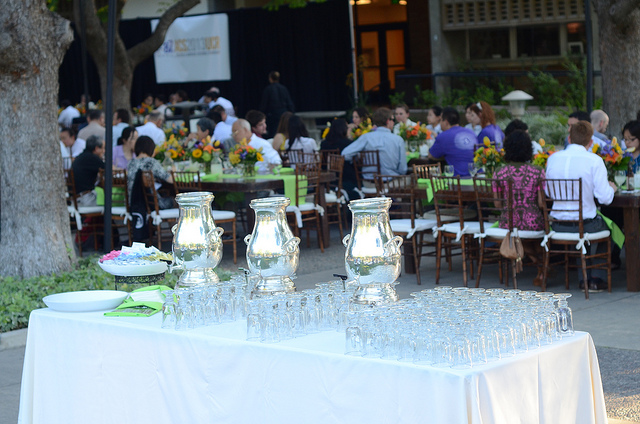} &
    \includegraphics[width=\segw,height=\segw]{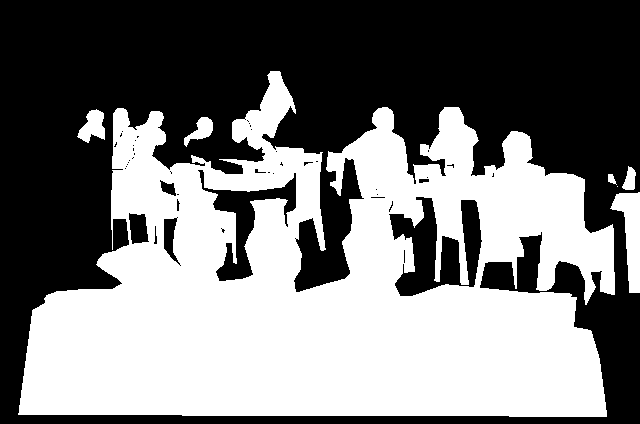} &
    \includegraphics[width=\segw,height=\segw]{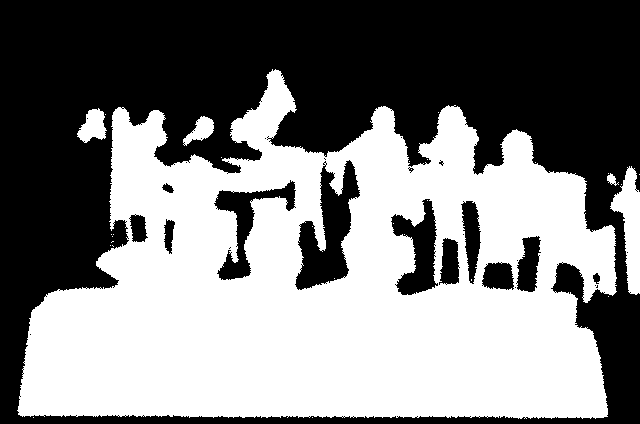} &
    \includegraphics[width=\segw,height=\segw]{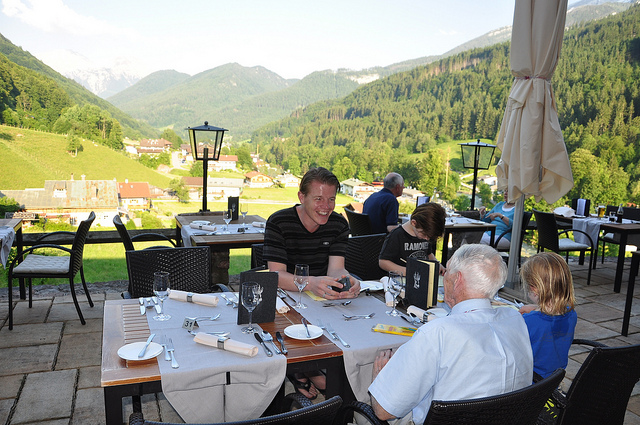} &
    \includegraphics[width=\segw,height=\segw]{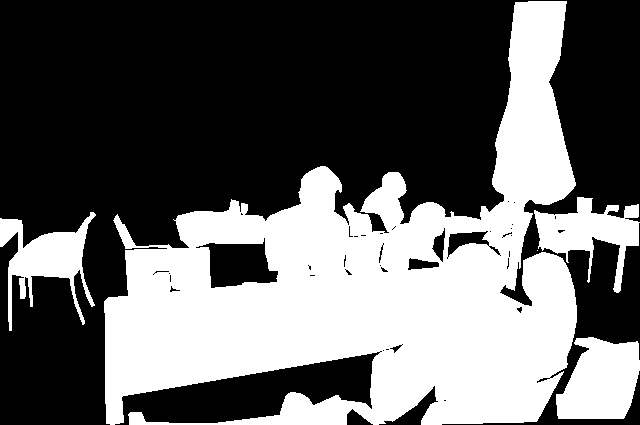} &
    \includegraphics[width=\segw,height=\segw]{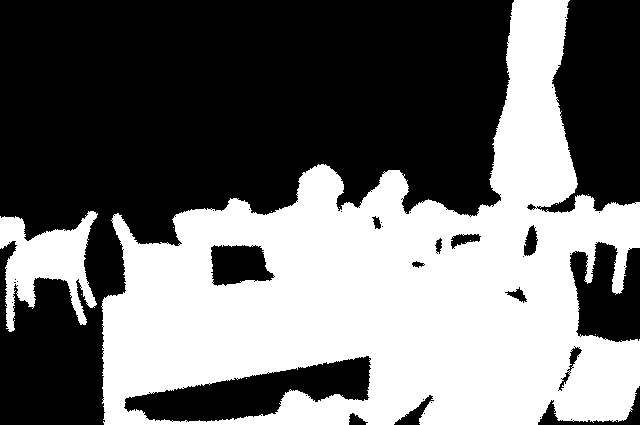} \\

  \end{tabular}

  \vspace{1mm}
  \hfill
  \makebox[0.30\textwidth][c]{%
    $\underbrace{\hspace{0.27\textwidth}}_{\text{\scriptsize Image $|$ GT $|$ Prediction}}$}%
  \hfill
  \makebox[0.30\textwidth][c]{%
    $\underbrace{\hspace{0.27\textwidth}}_{\text{\scriptsize Image $|$ GT $|$ Prediction}}$}%
  \hfill
  \makebox[0.30\textwidth][c]{%
    $\underbrace{\hspace{0.27\textwidth}}_{\text{\scriptsize Image $|$ GT $|$ Prediction}}$}%
  \hfill\null

  \vspace{1mm}
  \caption{%
    \textbf{Qualitative segmentation results of $\mathcal{H}_S$.}
    For each dataset (ADE20K, Kvasir-SEG, PASCAL VOC-2012, MS COCO-2014),
    three sample triplets are shown: input image, Ground Truth (GT),
    and CM-GLasso prediction (Ours).
    Our method produces precise boundaries and correctly captures
    long-range semantic dependencies
    (\emph{e.g.}, polyp edges, sky--water reflections,
    building--ground transitions), validating the benefit of
    joint ADMM optimization with cross-modal prior guidance.%
  }
  \label{fig:seg_vis}
\end{figure*}

\textbf{Interpretability:} (See Figure~\ref{fig:gam_vis}). Cross-attention mapping provides spatial transparency: reshaping $\mathbf{A}$'s rows reveals that nodes converge into semantic detectors (e.g., ``animal body''). Non-zero edges in $\boldsymbol{\Theta}^{(c)}$ explicitly link these physical regions, making $\mathcal{H}_C$'s partial correlation likelihoods traceable. For segmentation (Figure~\ref{fig:seg_vis}), Standard GLasso propagates noise via spurious edges, while Two-Stage accumulation breaks long-range ties. CM-GLasso uniquely preserves authentic remote pathways (e.g., sky-water reflections).

\textbf{Complexity:} Table~\ref{tab:complexity} merges runtime and theoretical complexity. The offline bottleneck is ADMM's $\mathcal{O}(p^3)$ eigendecomposition ($\sim$10 mins at $p=50$). Inference is highly efficient, needing only $\mathcal{O}(pN_p)$ for graph back-projection, easily supporting real-time pipelines.

\begin{table}[htbp]
\centering
\caption{Complexity \& Runtime (single NVIDIA A800).}
\label{tab:complexity}
\small
\begin{tabular}{@{}lcc@{}}
\toprule
\textbf{Module} & \textbf{Time (sec)} & \textbf{Complexity} \\
\midrule
ViT Feat. Extraction & 242 & $\mathcal{O}(NLd^2)$ \\
Cross-Attn \& Prior & 58 & $\mathcal{O}(NpN_pd_k + Cp^2N_p)$ \\
eBIC $k^*$ Selection & 268 & $\mathcal{O}(C|\mathcal{K}|p^3)$ \\
Joint ADMM Opt. & 7.2 & $\mathcal{O}(TCp^3)$ \\
$\mathcal{H}_C$/$\mathcal{H}_S$ Inference & 6.8 & $\mathcal{O}(Cp^2 + pN_p)$ \\
\bottomrule
\end{tabular}
\end{table}

\section{Conclusion}
\vspace{-1mm}

We introduced CM-GLasso, a unified topology-aware framework bridging deep representation learning and statistical graphical models. By integrating a \textbf{text visualization} strategy with a \textbf{unified SigLIP 2 encoder}, we resolve cross-modal feature inconsistencies. The proposed \textbf{cross-attention distillation} condenses high-dimensional patches into interpretable graph nodes, yielding spatially-aligned priors.

Our core innovation lies in the \textbf{Joint ADMM Optimization} which unifies tailored GLasso and common-specific structure learning (CSSL) into a single objective. By employing a \textbf{decoupled proxy supervision} strategy, we bypass the numerical instabilities of unrolled optimization while ensuring mathematically rigorous topology disentanglement. Results across eight benchmarks demonstrate that explicit probabilistic semantic structures serve as powerful inductive biases, significantly enhancing both discriminative and dense prediction tasks. Future work will explore the extension of this spatially-aware prior mechanism to temporal domains for video understanding.

\subsection{Limitations}

While CM-GLasso demonstrates superior performance across multiple benchmarks, it presents notable computational constraints regarding large-scale expansibility. Specifically, the offline ADMM optimization necessitates exact matrix eigenvalue decompositions at each iterative step, resulting in an $\mathcal{O}(TCp^3)$ computational complexity. Although this is highly efficient and easily tractable for moderate dataset settings with tens or hundreds of categories (e.g., CIFAR-100, CUB-200-2011), scaling this exact optimization framework to massive label spaces encompassing thousands of categories (e.g., ImageNet) introduces a linear computational bottleneck with respect to the number of classes $C$. Addressing this scaling challenge via low-rank matrix approximations or hierarchical category clustering remains a critical direction for future investigation.

\FloatBarrier
\bibliographystyle{plain}
\bibliography{reference}
\end{document}